\documentclass[10pt,twocolumn,letterpaper]{article}

\usepackage[pagenumbers]{cvpr} 
\usepackage{amsthm}
\newtheorem{proposition}{Proposition}
\usepackage{array}
\usepackage{multirow}
\usepackage{colortbl}
\usepackage{needspace}
\usepackage{tikz}

\definecolor{cvprblue}{rgb}{0.21,0.49,0.74}
\usepackage[pagebackref,breaklinks,colorlinks,allcolors=cvprblue]{hyperref}

\def\confName{CVPR}
\def\confYear{2027}

\title{Policy Improvement–Guided Replay for Continual Reinforcement Learning}

\author{Qin-Wen Luo\thanks{The authors contribute equally.}, 
Haoxuan Gao\footnotemark[1], 
Sheng Ren, 
Ming-Kun Xie, 
Xiang Chen\thanks{Corresponding Author.}, 
Sheng-Jun Huang\footnotemark[2]\\
Nanjing University of Aeronautics and Astronautics\\
Nanjing, China\\
}
\begin{document}
\raggedbottom
\def\CoLiRPrepareOnly{1}
\ifdefined\CoLiRBackgroundTable
\else
\usetikzlibrary{backgrounds,calc}
\definecolor{CoLiRTableTint}{RGB}{236,242,247}
\newsavebox{\CoLiRTableBox}
\newlength{\CoLiRTableRowHeight}
\newcommand{\CoLiRBackgroundRect}[4]{%
  \edef\CoLiRXA{\the\dimexpr#1\relax}%
  \edef\CoLiRYA{\the\dimexpr#2+0.3pt\relax}%
  \edef\CoLiRXB{\the\dimexpr#3\relax}%
  \edef\CoLiRYB{\the\dimexpr#4-0.3pt\relax}%
  \ifnum\CoLiRBackgroundType=0
    \edef\CoLiRYA{\the\dimexpr\CoLiRYA+1.7pt\relax}%
  \fi
  \ifnum\CoLiRBackgroundType=1
    \edef\CoLiRYB{\the\dimexpr\CoLiRYB-2.9pt\relax}%
  \fi
  \ifnum\CoLiRBackgroundType=2
    \edef\CoLiRYB{\the\dimexpr\CoLiRYB-2.9pt\relax}%
  \fi
  \fill[CoLiRTableTint]
    ($(colir-table.south west)+(\CoLiRXA,\CoLiRYA)$)
    rectangle ($(colir-table.south west)+(\CoLiRXB,\CoLiRYB)$);%
  \begin{scope}
    \clip ($(colir-table.south west)+(\CoLiRXA,\CoLiRYA)$) rectangle ($(colir-table.south west)+(\CoLiRXB,\CoLiRYB)$);
    \ifnum\CoLiRBackgroundType=0
      \foreach \CoLiRGapX in {.065\textwidth-3\arrayrulewidth-4pt,.2\textwidth-3\arrayrulewidth,.7\textwidth-3\arrayrulewidth}{
        \edef\CoLiRGapPos{\the\dimexpr\CoLiRGapX\relax}
        \draw[white,line width=2.8pt] ($(colir-table.south west)+(\CoLiRGapPos,\CoLiRYA)$) -- ($(colir-table.south west)+(\CoLiRGapPos,\CoLiRYB)$);
      }
    \fi
    \ifnum\CoLiRBackgroundType=3
      \edef\CoLiRGapY{\the\dimexpr\heavyrulewidth+2\aboverulesep+\CoLiRTableRowHeight+.5\lightrulewidth\relax}
      \draw[white,line width=2.8pt] ($(colir-table.south west)+(\CoLiRXA,\CoLiRGapY)$) -- ($(colir-table.south west)+(\CoLiRXB,\CoLiRGapY)$);
      \edef\CoLiRGapY{\the\dimexpr\ht\CoLiRTableBox+\dp\CoLiRTableBox-\heavyrulewidth-\belowrulesep-2\CoLiRTableRowHeight-\cmidrulewidth-2\aboverulesep-\belowrulesep-.5\lightrulewidth\relax}
      \draw[white,line width=2.8pt] ($(colir-table.south west)+(\CoLiRXA,\CoLiRGapY)$) -- ($(colir-table.south west)+(\CoLiRXB,\CoLiRGapY)$);
    \fi
  \end{scope}

}
\newenvironment{CoLiRBackgroundTable}[1]{%
  \def\CoLiRBackgroundType{#1}%
  \setlength{\CoLiRTableRowHeight}{\arraystretch\dimexpr\ht\strutbox+\dp\strutbox\relax}%
  \begin{lrbox}{\CoLiRTableBox}%
}{%
  \end{lrbox}%
  \begin{tikzpicture}[baseline=(colir-table.base)]
    \node[inner sep=0pt,outer sep=0pt] (colir-table) {\usebox{\CoLiRTableBox}};
    \begin{scope}[on background layer]
      \ifnum\CoLiRBackgroundType=0
        \CoLiRBackgroundRect{.065\textwidth-3\arrayrulewidth-4pt}
          {\heavyrulewidth}{\wd\CoLiRTableBox}
          {\heavyrulewidth+\aboverulesep+\CoLiRTableRowHeight}
        \CoLiRBackgroundRect{.065\textwidth-3\arrayrulewidth-4pt}
          {\heavyrulewidth+9\CoLiRTableRowHeight+\cmidrulewidth+\lightrulewidth+2\aboverulesep+2\belowrulesep}
          {\wd\CoLiRTableBox}
          {\heavyrulewidth+10\CoLiRTableRowHeight+\cmidrulewidth+\lightrulewidth+3\aboverulesep+2\belowrulesep}
        \CoLiRBackgroundRect{.065\textwidth-3\arrayrulewidth-4pt}
          {\heavyrulewidth+18\CoLiRTableRowHeight+2\cmidrulewidth+2\lightrulewidth+4\aboverulesep+4\belowrulesep}
          {\wd\CoLiRTableBox}
          {\heavyrulewidth+19\CoLiRTableRowHeight+2\cmidrulewidth+2\lightrulewidth+5\aboverulesep+4\belowrulesep}
      \fi
      \ifnum\CoLiRBackgroundType=1
        \CoLiRBackgroundRect{0pt}{\heavyrulewidth+\aboverulesep+3\CoLiRTableRowHeight}
          {\wd\CoLiRTableBox}{\heavyrulewidth+\aboverulesep+4\CoLiRTableRowHeight+\belowrulesep}
      \fi
      \ifnum\CoLiRBackgroundType=2
        \CoLiRBackgroundRect{0pt}{\heavyrulewidth+\aboverulesep+4\CoLiRTableRowHeight}
          {\wd\CoLiRTableBox}{\heavyrulewidth+\aboverulesep+5\CoLiRTableRowHeight+\belowrulesep}
      \fi
      \ifnum\CoLiRBackgroundType=3
        \CoLiRBackgroundRect{\wd\CoLiRTableBox-.08\textwidth}{\heavyrulewidth}
          {\wd\CoLiRTableBox}
          {\heavyrulewidth+13\CoLiRTableRowHeight+2\lightrulewidth+4\cmidrulewidth+7\aboverulesep+7\belowrulesep}
      \fi
    \end{scope}
    \ifnum\CoLiRBackgroundType=3
      \edef\CoLiRXPos{\the\dimexpr.38\textwidth\relax}%
      \edef\CoLiRYLow{\the\dimexpr\heavyrulewidth+\aboverulesep+\CoLiRTableRowHeight+\aboverulesep+\lightrulewidth+\belowrulesep\relax}%
      \edef\CoLiRYHigh{\the\dimexpr\ht\CoLiRTableBox+\dp\CoLiRTableBox-\heavyrulewidth-\belowrulesep-2\CoLiRTableRowHeight-\cmidrulewidth-\aboverulesep-\belowrulesep-\aboverulesep-\lightrulewidth-\belowrulesep\relax}%
      \draw[line width=\arrayrulewidth] ($(colir-table.south west)+(\CoLiRXPos,\CoLiRYLow)$) -- ($(colir-table.south west)+(\CoLiRXPos,\CoLiRYHigh)$);
    \fi
    \ifnum\CoLiRBackgroundType=0
      \foreach \CoLiRX [count=\CoLiRLine] in {.065\textwidth-3\arrayrulewidth-4pt,.2\textwidth-3\arrayrulewidth,.7\textwidth-3\arrayrulewidth}{%
        \edef\CoLiRXPos{\the\dimexpr\CoLiRX\relax}%
        \foreach \CoLiRBlock in {0,1,2}{%
          \edef\CoLiRYLow{\the\dimexpr\heavyrulewidth+\aboverulesep+\CoLiRBlock\dimexpr9\CoLiRTableRowHeight+\cmidrulewidth+\lightrulewidth+2\aboverulesep+2\belowrulesep\relax\relax}%
          \edef\CoLiRYHigh{\the\dimexpr\CoLiRYLow+9\CoLiRTableRowHeight+\cmidrulewidth+\aboverulesep+\belowrulesep\relax}%
          \ifnum\CoLiRLine=1
            \draw[line width=\arrayrulewidth] ($(colir-table.south west)+(\CoLiRXPos,\CoLiRYLow)$) -- ($(colir-table.south west)+(\CoLiRXPos,\CoLiRYHigh)$);
          \else
          \edef\CoLiRYLowerEnd{\the\dimexpr\CoLiRYLow+5\CoLiRTableRowHeight\relax}%
          \edef\CoLiRYUpperStart{\the\dimexpr\CoLiRYLowerEnd+\aboverulesep+\cmidrulewidth+\belowrulesep\relax}%
          \draw[line width=\arrayrulewidth] ($(colir-table.south west)+(\CoLiRXPos,\CoLiRYLow)$) -- ($(colir-table.south west)+(\CoLiRXPos,\CoLiRYLowerEnd)$);
          \draw[line width=\arrayrulewidth] ($(colir-table.south west)+(\CoLiRXPos,\CoLiRYUpperStart)$) -- ($(colir-table.south west)+(\CoLiRXPos,\CoLiRYHigh)$);
          \fi
        }

      }
    \fi
  \end{tikzpicture}%
}
\fi

\captionsetup[table]{position=top,skip=5pt,labelfont=bf,justification=raggedright,singlelinecheck=false}
\setlength{\heavyrulewidth}{0.7pt}
\setlength{\lightrulewidth}{0.4pt}
\setlength{\cmidrulewidth}{0.3pt}
\setlength{\aboverulesep}{.4ex}
\setlength{\belowrulesep}{.65ex}
\newcommand{\ExpMainCaption}{Performance comparison of different methods on the MRCL benchmark at three model scales of Qwen3-VL. The best results within each scale are shown in \textbf{bold}. The second-best results are \underline{underlined}.}
\newcommand{\ExpOODCaption}{Performance comparison on out-of-distribution benchmarks after sequential training. The best results among trained models are shown in \textbf{bold}. The second-best results are \underline{underlined}.}
\newcommand{\ExpFilteringCaption}{Ablation results for different components.}
\newcommand{\ExpScoreCaption}{Performance comparison under different replay selection methods.}
\newcommand{\ExpKLCaption}{Performance with and without replay KL regularization on Qwen3-VL-4B.}
\newcommand{\ExpMemoryCaption}{Performance under different replay memory capacities.}
\newcommand{\ExpScatterCaption}{Joint distributions of correct-response log-likelihood increase and incorrect-response log-likelihood decrease across T2--T5.}
\newcommand{\ExpHeatmapCaption}{Replay selection rates across different combinations of early and late correct-response counts for T2--T5.}
\newcommand{\CoLiRMainResultsTable}{%
\begin{table*}[!t]
\centering
\caption{\ExpMainCaption}
\label{tab:main-results}
\begingroup
\small
\setlength{\tabcolsep}{2.2pt}
\renewcommand{\arraystretch}{1.08}
\begin{CoLiRBackgroundTable}{0}
\begin{tabular}{>{\raggedright\arraybackslash}p{\dimexpr.065\textwidth-2\tabcolsep-3\arrayrulewidth\relax}>{\raggedright\arraybackslash}p{\dimexpr.135\textwidth-2\tabcolsep\relax}*{5}{>{\centering\arraybackslash}p{\dimexpr.1\textwidth-2\tabcolsep\relax}}*{3}{>{\centering\arraybackslash}p{\dimexpr.1\textwidth-2\tabcolsep\relax}}}
\toprule
\multicolumn{1}{l}{\multirow{2}{*}{\textbf{Size}}} & \multirow{2}{*}{\textbf{Method}} & \multicolumn{5}{c}{\textbf{Final accuracy on each task (\%)}} & \multicolumn{3}{c}{\textbf{Aggregated accuracy (\%)}} \\
\cmidrule(lr){3-7}\cmidrule(l){8-10}
 & & \textbf{MedBook} & \textbf{Navig.} & \textbf{We-Math2} & \textbf{Puzzle} & \textbf{FinMME} & \textbf{MFN$\uparrow$} & \textbf{MTA$\uparrow$} & \textbf{MFT$\uparrow$} \\
\midrule

 & FullSFT & 45.67 & 19.11 & 33.90 & \underline{84.00} & 32.69 & 43.07 & 51.97 & \textbf{64.33} \\
 & LoRA & \underline{59.08} & \underline{49.17} & 35.75 & \textbf{88.00} & 32.80 & \underline{52.96} & \underline{56.87} & 62.34 \\
 & O-LoRA & 57.72 & 22.82 & 35.22 & 75.50 & 35.17 & 45.29 & 51.32 & 57.66 \\
 & RegLoRA & \textbf{61.63} & 43.04 & 35.51 & 81.25 & 33.00 & 50.89 & 54.41 & 57.80 \\
\cmidrule(lr){2-10}
 & GRPO & 50.93 & 19.11 & 45.14 & 69.50 & 41.78 & 45.29 & 49.88 & 55.17 \\
 & GRPO$^{*}$ & 52.46 & 19.48 & \textbf{48.44} & 55.25 & 41.27 & 43.38 & 44.91 & 45.87 \\
 & GSPO & 47.03 & 13.36 & 40.24 & 76.75 & \textbf{43.14} & 44.10 & 53.19 & 63.43 \\
 & CPO & 56.54 & 33.58 & \underline{47.34} & 75.75 & 42.33 & 51.11 & 54.17 & 61.31 \\
\multirow{-9}{*}{\textbf{2B}} & \textbf{CoLiR (Ours)} & 53.14 & \textbf{52.69} & 45.80 & 82.50 & \underline{42.48} & \textbf{55.32} & \textbf{59.84} & \underline{63.68} \\
\midrule

 & FullSFT & 53.48 & 24.86 & 37.78 & 79.75 & 35.42 & 46.26 & 57.14 & 66.35 \\
 & LoRA & 66.89 & 45.64 & 42.26 & 35.00 & 41.62 & 46.28 & 44.93 & 51.94 \\
 & O-LoRA & \underline{75.04} & 49.54 & 43.67 & \textbf{93.75} & 41.32 & 60.66 & 63.22 & 66.97 \\
 & RegLoRA & \textbf{76.91} & \underline{74.95} & 44.70 & 37.75 & 41.02 & 55.07 & 56.47 & 59.09 \\
\cmidrule(lr){2-10}
 & GRPO & 60.95 & 18.74 & 51.54 & 76.00 & 43.14 & 50.07 & 55.60 & 62.48 \\
 & GRPO$^{*}$ & 48.90 & 2.04 & 53.18 & 28.25 & 41.17 & 34.71 & 35.44 & 36.54 \\
 & GSPO & 58.57 & 35.62 & 51.60 & \underline{91.50} & \underline{45.56} & 56.57 & 64.87 & \textbf{73.45} \\
 & CPO & 60.27 & 60.11 & \underline{55.30} & 83.00 & \textbf{49.29} & \underline{61.59} & \underline{66.95} & 70.78 \\
\multirow{-9}{*}{\textbf{4B}} & \textbf{CoLiR (Ours)} & 62.82 & \textbf{91.28} & \textbf{57.96} & 89.50 & 45.21 & \textbf{69.35} & \textbf{70.86} & \underline{72.27} \\
\midrule

 & FullSFT & 58.57 & 25.60 & 43.20 & 35.75 & 37.79 & 40.18 & 49.26 & 58.24 \\
 & LoRA & 65.87 & 47.12 & 44.89 & \textbf{95.25} & 42.28 & \underline{59.08} & 61.23 & 70.85 \\
 & O-LoRA & \underline{72.50} & 52.32 & 44.78 & 35.25 & 42.79 & 49.53 & 50.37 & 53.75 \\
 & RegLoRA & \textbf{72.84} & \underline{72.36} & 46.39 & 55.50 & 44.70 & 58.36 & 60.67 & 62.93 \\
\cmidrule(lr){2-10}
 & GRPO & 62.14 & 3.90 & 56.35 & 67.00 & 45.16 & 46.91 & 49.90 & 52.98 \\
 & GRPO$^{*}$ & 67.06 & 27.27 & \textbf{62.10} & 58.50 & \textbf{48.59} & 52.70 & 55.92 & 59.86 \\
 & GSPO & 58.74 & 3.15 & 56.37 & \underline{90.25} & 45.46 & 50.79 & 60.85 & \textbf{72.71} \\
 & CPO & 64.52 & 33.58 & \underline{56.43} & 89.25 & \underline{47.12} & 58.18 & \underline{65.49} & \underline{72.51} \\
\multirow{-9}{*}{\textbf{8B}} & \textbf{CoLiR (Ours)} & 60.61 & \textbf{94.81} & 56.37 & 84.00 & 45.06 & \textbf{68.17} & \textbf{70.36} & 72.44 \\
\bottomrule
\end{tabular}
\end{CoLiRBackgroundTable}
\endgroup

\end{table*}
}
\newcommand{\CoLiRMainResultsInline}{%
\begin{minipage}{\textwidth}
\captionsetup{type=table}
\centering
\caption{\ExpMainCaption}
\label{tab:main-results}
\begingroup
\small
\setlength{\tabcolsep}{2.2pt}
\renewcommand{\arraystretch}{1.08}
\begin{CoLiRBackgroundTable}{0}
\begin{tabular}{>{\raggedright\arraybackslash}p{\dimexpr.065\textwidth-2\tabcolsep-3\arrayrulewidth\relax}>{\raggedright\arraybackslash}p{\dimexpr.135\textwidth-2\tabcolsep\relax}*{5}{>{\centering\arraybackslash}p{\dimexpr.1\textwidth-2\tabcolsep\relax}}*{3}{>{\centering\arraybackslash}p{\dimexpr.1\textwidth-2\tabcolsep\relax}}}
\toprule
\multicolumn{1}{l}{\multirow{2}{*}{\textbf{Size}}} & \multirow{2}{*}{\textbf{Method}} & \multicolumn{5}{c}{\textbf{Final accuracy on each task (\%)}} & \multicolumn{3}{c}{\textbf{Aggregated accuracy (\%)}} \\
\cmidrule(lr){3-7}\cmidrule(l){8-10}
 & & \textbf{MedBook} & \textbf{Navig.} & \textbf{We-Math2} & \textbf{Puzzle} & \textbf{FinMME} & \textbf{MFN$\uparrow$} & \textbf{MTA$\uparrow$} & \textbf{MFT$\uparrow$} \\
\midrule

 & FullSFT & 45.67 & 19.11 & 33.90 & \underline{84.00} & 32.69 & 43.07 & 51.97 & \textbf{64.33} \\
 & LoRA & \underline{59.08} & \underline{49.17} & 35.75 & \textbf{88.00} & 32.80 & \underline{52.96} & \underline{56.87} & 62.34 \\
 & O-LoRA & 57.72 & 22.82 & 35.22 & 75.50 & 35.17 & 45.29 & 51.32 & 57.66 \\
 & RegLoRA & \textbf{61.63} & 43.04 & 35.51 & 81.25 & 33.00 & 50.89 & 54.41 & 57.80 \\
\cmidrule(lr){2-10}
 & GRPO & 50.93 & 19.11 & 45.14 & 69.50 & 41.78 & 45.29 & 49.88 & 55.17 \\
 & GRPO$^{*}$ & 52.46 & 19.48 & \textbf{48.44} & 55.25 & 41.27 & 43.38 & 44.91 & 45.87 \\
 & GSPO & 47.03 & 13.36 & 40.24 & 76.75 & \textbf{43.14} & 44.10 & 53.19 & 63.43 \\
 & CPO & 56.54 & 33.58 & \underline{47.34} & 75.75 & 42.33 & 51.11 & 54.17 & 61.31 \\
\multirow{-9}{*}{\textbf{2B}} & \textbf{CoLiR (Ours)} & 53.14 & \textbf{52.69} & 45.80 & 82.50 & \underline{42.48} & \textbf{55.32} & \textbf{59.84} & \underline{63.68} \\
\midrule

 & FullSFT & 53.48 & 24.86 & 37.78 & 79.75 & 35.42 & 46.26 & 57.14 & 66.35 \\
 & LoRA & 66.89 & 45.64 & 42.26 & 35.00 & 41.62 & 46.28 & 44.93 & 51.94 \\
 & O-LoRA & \underline{75.04} & 49.54 & 43.67 & \textbf{93.75} & 41.32 & 60.66 & 63.22 & 66.97 \\
 & RegLoRA & \textbf{76.91} & \underline{74.95} & 44.70 & 37.75 & 41.02 & 55.07 & 56.47 & 59.09 \\
\cmidrule(lr){2-10}
 & GRPO & 60.95 & 18.74 & 51.54 & 76.00 & 43.14 & 50.07 & 55.60 & 62.48 \\
 & GRPO$^{*}$ & 48.90 & 2.04 & 53.18 & 28.25 & 41.17 & 34.71 & 35.44 & 36.54 \\
 & GSPO & 58.57 & 35.62 & 51.60 & \underline{91.50} & \underline{45.56} & 56.57 & 64.87 & \textbf{73.45} \\
 & CPO & 60.27 & 60.11 & \underline{55.30} & 83.00 & \textbf{49.29} & \underline{61.59} & \underline{66.95} & 70.78 \\
\multirow{-9}{*}{\textbf{4B}} & \textbf{CoLiR (Ours)} & 62.82 & \textbf{91.28} & \textbf{57.96} & 89.50 & 45.21 & \textbf{69.35} & \textbf{70.86} & \underline{72.27} \\
\midrule

 & FullSFT & 58.57 & 25.60 & 43.20 & 35.75 & 37.79 & 40.18 & 49.26 & 58.24 \\
 & LoRA & 65.87 & 47.12 & 44.89 & \textbf{95.25} & 42.28 & \underline{59.08} & 61.23 & 70.85 \\
 & O-LoRA & \underline{72.50} & 52.32 & 44.78 & 35.25 & 42.79 & 49.53 & 50.37 & 53.75 \\
 & RegLoRA & \textbf{72.84} & \underline{72.36} & 46.39 & 55.50 & 44.70 & 58.36 & 60.67 & 62.93 \\
\cmidrule(lr){2-10}
 & GRPO & 62.14 & 3.90 & 56.35 & 67.00 & 45.16 & 46.91 & 49.90 & 52.98 \\
 & GRPO$^{*}$ & 67.06 & 27.27 & \textbf{62.10} & 58.50 & \textbf{48.59} & 52.70 & 55.92 & 59.86 \\
 & GSPO & 58.74 & 3.15 & 56.37 & \underline{90.25} & 45.46 & 50.79 & 60.85 & \textbf{72.71} \\
 & CPO & 64.52 & 33.58 & \underline{56.43} & 89.25 & \underline{47.12} & 58.18 & \underline{65.49} & \underline{72.51} \\
\multirow{-9}{*}{\textbf{8B}} & \textbf{CoLiR (Ours)} & 60.61 & \textbf{94.81} & 56.37 & 84.00 & 45.06 & \textbf{68.17} & \textbf{70.36} & 72.44 \\
\bottomrule
\end{tabular}
\end{CoLiRBackgroundTable}
\endgroup

\end{minipage}
}
\newcommand{\CoLiROODTable}{%
\begin{table*}[t]
\centering
\caption{\ExpOODCaption}
\label{tab:ood-4b}
\begingroup
\small
\setlength{\tabcolsep}{2.0pt}
\renewcommand{\arraystretch}{1.15}
\makebox[\textwidth][c]{%
\begin{CoLiRBackgroundTable}{3}
\begin{tabular}{>{\centering\arraybackslash}p{\dimexpr.14\textwidth-2\tabcolsep\relax}>{\centering\arraybackslash}p{\dimexpr.24\textwidth-2\tabcolsep\relax}*{6}{>{\centering\arraybackslash}p{\dimexpr.08\textwidth-2\tabcolsep\relax}}>{\centering\arraybackslash}p{\dimexpr.08\textwidth-2\tabcolsep\relax}}
\toprule
\multirow{2}{*}{\textbf{Dimension}} & \multirow{2}{*}{\textbf{Benchmark}} & \multicolumn{7}{c}{\textbf{Qwen3-VL-4B}} \\
\cmidrule(l){3-9}
 &  & \textbf{Base} & \textbf{FullSFT} & \textbf{GRPO} & \textbf{GRPO$^{*}$} & \textbf{GSPO} & \textbf{CPO} & \textbf{CoLiR} \\
\midrule
\multirow{4}{*}{STEM} & MMMU-Pro$_{\mathrm{full}}$~\cite{mmmupro} & 46.09 & 23.41 & 35.10 & \textbf{41.95} & 33.51 & 37.13 & \underline{41.50} \\
& MathVista$_{\mathrm{mini}}$~\cite{mathvista} & 55.80 & 48.40 & 58.50 & 64.20 & 61.50 & \underline{66.50} & \textbf{67.00} \\
& MathVision$_{\mathrm{TEST}}$~\cite{mathvision} & 21.22 & 16.78 & 26.28 & 23.88 & 23.55 & \underline{28.55} & \textbf{29.70} \\
& MathVerse$_{\mathrm{mini}}$~\cite{mathverse} & 36.90 & 27.13 & 40.46 & 40.56 & 42.94 & \textbf{48.15} & \underline{45.13} \\
\cmidrule(r){1-2}
\multirow{2}{*}{General VQA} & RealWorldQA~\cite{realworldqa} & 68.24 & 50.33 & \textbf{65.23} & 56.47 & \underline{61.83} & 60.39 & 57.52 \\
& MMStar~\cite{mmstar} & 59.13 & 53.00 & 58.80 & 57.13 & 60.53 & \underline{62.27} & \textbf{62.40} \\
\cmidrule(r){1-2}
Hallucination & POPE~\cite{pope} & 87.26 & 84.03 & 82.88 & \textbf{85.46} & \underline{85.09} & 84.71 & 83.89 \\
\cmidrule(r){1-2}
\multirow{3}{*}{\shortstack[c]{Chart/Doc\\Understanding}} & DocVQA$_{\mathrm{VAL}}$~\cite{docvqa} & 65.34 & 62.46 & 55.77 & \textbf{69.71} & 57.84 & \underline{69.13} & 60.01 \\
& CharXiv$_{\mathrm{DQ}}$~\cite{charxiv} & 71.86 & 42.41 & 43.98 & \textbf{73.00} & 49.10 & \underline{54.93} & 53.62 \\
& CharXiv$_{\mathrm{RQ}}$~\cite{charxiv} & 33.90 & 24.90 & 28.00 & 25.80 & 30.10 & \textbf{33.30} & \underline{31.30} \\
\cmidrule(r){1-2}
Counting & CountBenchQA~\cite{countbench,paligemma} & 84.32 & 69.25 & 60.69 & \underline{71.08} & 67.62 & 58.66 & \textbf{78.62} \\
\midrule
\multicolumn{2}{c}{\textbf{Average}} & 57.28 & 45.65 & 50.52 & \underline{55.39} & 52.15 & 54.88 & \textbf{55.52} \\
\bottomrule
\end{tabular}
\end{CoLiRBackgroundTable}%
}
\endgroup

\end{table*}
}
\newcommand{\CoLiRScatterFigure}{%
\begin{figure*}[t]
\centering
\includegraphics[width=0.94\textwidth]{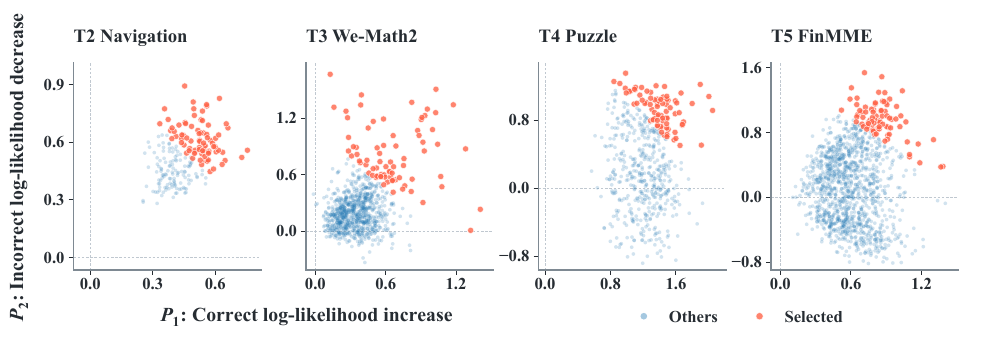}
\caption{\ExpScatterCaption}
\label{fig:likelihood-components}
\end{figure*}
}
\newcommand{\CoLiRHeatmapFigure}{%
\begin{figure*}[t]
\centering
\includegraphics[width=0.94\textwidth,trim=0 7bp 0 7bp,clip]{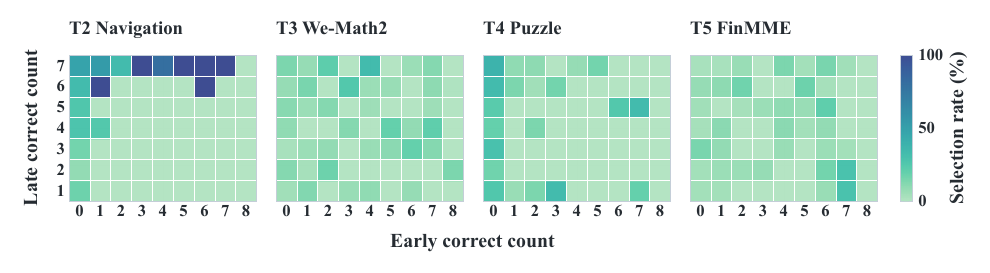}
\caption{\ExpHeatmapCaption}
\label{fig:selection-rate}
\end{figure*}
}
\newcommand{\CoLiRExperiments}{%

\section{Experiments}
\label{sec:experiment}
\CoLiRScatterFigure
\CoLiRHeatmapFigure

\subsection{Experimental Setup}
\label{sec:experimental-setup}

\paragraph{Datasets.}
We conduct CRL experiments on the MRCL benchmark~\cite{cpo}, following the task sequence MedBookVQA, Navigation, We-Math2, Puzzle, and FinMME. These tasks cover medical visual question answering, spatial navigation, mathematical reasoning, visual puzzles, and financial understanding. 
Due to answer leakage in the dataset provided by CPO~\cite{cpo}, we correct formatting inconsistencies, answer annotations, and scoring errors, and remove invalid or duplicate examples from the original benchmark. 
We evaluate final performance, performance throughout training, and new-task learning using Mean Final Accuracy (MFN), Mean Task Accuracy (MTA), and Mean Finetune Accuracy (MFT), respectively. 
We also assess general capability retention on out-of-distribution (OOD) benchmarks covering STEM reasoning, general visual question answering, hallucination, chart and document understanding, and counting.

\paragraph{Experimental settings.}
We conduct experiments on Qwen3-VL at three scales: 2B, 4B, and 8B. CoLiR uses GSPO~\cite{gspo} with historical replay. By ranking prompts using the contrastive score, CoLiR retains at most $B=80$ prompts per historical task, with $B$ much smaller than the size of each task's training set. We jointly optimize performance on current and historical tasks using responses generated by the current policy, with a training batch size of 32 prompts; each batch includes 4 replay prompts at T2 and 8 at T3--T5, sampled uniformly across historical tasks. The replay KL loss coefficient is $\beta=0.01$, with the reference policy frozen at the start of the current task. To evaluate CoLiR's effectiveness, we compare CoLiR with supervised baselines, including FullSFT, LoRA~\cite{lora}, O-LoRA~\cite{olora}, and RegLoRA~\cite{reglora}, and RL baselines, including GRPO~\cite{grpo}, GRPO$^{*}$~\cite{grpo} (with a KL coefficient of 0.04), GSPO~\cite{gspo}, and CPO~\cite{cpo}. CPO employs parameter regularization, a common technique in continual learning, to limit changes to parameters associated with historical tasks. After correcting the MRCL data and scoring issues, we reproduce all baselines using their official code. Further baseline details are provided in Appendix B.2.

\subsection{Main Results}
\label{sec:main-results}

\textbf{CoLiR consistently improves task retention across model scales.}
CoLiR achieves the highest MFN and MTA at all three model scales (Table~\ref{tab:main-results}).
Consistent with CPO's findings~\cite{cpo}, the MFT--MFN gap for GRPO and GSPO indicates forgetting during subsequent task learning.
CoLiR substantially improves retention relative to these baselines, and its MFN advantage over CPO increases with model size, reaching 9.99 percentage points at 8B.
Despite allocating part of the training budget to replay, CoLiR achieves MFT comparable to that of GSPO across all three model sizes, and outperforms CPO at 2B and 4B.
CoLiR also maintains high Navigation accuracy after subsequent tasks and shows improvements across tasks with different reasoning requirements at 4B.
Together, these results demonstrate that CoLiR effectively mitigates forgetting while preserving new-task learning ability.

\begin{table*}[ht]
\noindent
\begin{minipage}[t]{\columnwidth}
\vspace{0pt}\captionsetup{type=table}
\centering
\caption{\ExpFilteringCaption}
\label{tab:ablation-filtering}
\begingroup
\small
\setlength{\tabcolsep}{3pt}
\renewcommand{\arraystretch}{1.10}
\makebox[\linewidth][c]{%
\begin{CoLiRBackgroundTable}{1}
\begin{tabular}{>{\raggedright\arraybackslash}p{\dimexpr.40\linewidth-2\tabcolsep\relax}>{\centering\arraybackslash}p{\dimexpr.15\linewidth-2\tabcolsep\relax}>{\centering\arraybackslash}p{\dimexpr.15\linewidth-2\tabcolsep\relax}>{\centering\arraybackslash}p{\dimexpr.15\linewidth-2\tabcolsep\relax}}
\toprule
\textbf{Method} & \textbf{MFN$\uparrow$} & \textbf{MTA$\uparrow$} & \textbf{MFT$\uparrow$} \\
\midrule
CoLiR & \textbf{69.35} & \textbf{70.86} & \textbf{72.27} \\
w/o accuracy filtering & 60.78 & 63.51 & 70.60 \\
w/o contrastive selection & 66.59 & 68.77 & 71.10 \\
w/o replay KL & 66.99 & 68.16 & 70.21 \\
\bottomrule
\end{tabular}
\end{CoLiRBackgroundTable}%
}
\endgroup

\end{minipage}\hfill\begin{minipage}[t]{\columnwidth}
\vspace{0pt}\captionsetup{type=table}
\centering
\caption{\ExpScoreCaption}
\label{tab:ablation-score}
\begingroup
\small
\setlength{\tabcolsep}{3pt}
\renewcommand{\arraystretch}{1.10}
\makebox[\linewidth][c]{%
\begin{CoLiRBackgroundTable}{2}
\begin{tabular}{>{\raggedright\arraybackslash}p{\dimexpr.40\linewidth-2\tabcolsep\relax}>{\centering\arraybackslash}p{\dimexpr.15\linewidth-2\tabcolsep\relax}>{\centering\arraybackslash}p{\dimexpr.15\linewidth-2\tabcolsep\relax}>{\centering\arraybackslash}p{\dimexpr.15\linewidth-2\tabcolsep\relax}}
\toprule
\textbf{Selection method} & \textbf{MFN$\uparrow$} & \textbf{MTA$\uparrow$} & \textbf{MFT$\uparrow$} \\
\midrule
Contrastive (CoLiR) & \textbf{69.35} & \textbf{70.86} & \textbf{72.27} \\
Accuracy gain & 63.08 & 65.44 & 70.06 \\
$P_1$ only & 66.80 & 68.50 & 70.64 \\
$P_2$ only & 64.00 & 67.25 & 69.80 \\
Random & 66.59 & 68.77 & 71.10 \\
\bottomrule
\end{tabular}
\end{CoLiRBackgroundTable}%
}
\endgroup

\end{minipage}
\end{table*}

\subsection{General Capability Retention}
\label{sec:general-retention}

\textbf{CoLiR preserves general capabilities while learning new tasks.} Although CPO and GRPO$^{*}$ achieve average OOD scores similar to CoLiR (Table~\ref{tab:ood-4b}), they rely on explicit parameter regularization and a stronger KL constraint applied to all training prompts, respectively, both of which restrict model updates. CoLiR instead emphasizes RL optimization on current-task prompts and selected historical prompts, with a smaller KL coefficient applied only to replayed prompts. Taken together with the results in Table~\ref{tab:main-results}, these findings show that CoLiR achieves comparable general capability preservation while attaining stronger task acquisition and retention than both methods at 4B. Using the same GSPO optimizer, CoLiR retains substantially more of the original model's general capabilities than vanilla GSPO across multiple evaluation dimensions. This contrast supports the contribution of valuable historical replay, together with its local KL constraint, to capability preservation. Overall, CoLiR improves performance on the task sequence while limiting degradation in general capabilities.

\subsection{Ablation Studies}
\label{sec:ablation-studies}
\textit{Component Ablations.} \textbf{Filtering, contrastive selection, and replay KL jointly improve performance.} As shown in Table~\ref{tab:ablation-filtering}, the full method performs best across all three metrics. Removing candidate filtering produces the largest decrease in final accuracy, while only random sampling also weakens retention. These results support two-stage selection: filtering defines the eligible pool, and contrastive scoring prioritizes prompts whose likelihood changes indicate substantial policy improvement. In addition, removing replay KL lowers all three metrics, affecting both task acquisition and retention.
These results support the empirical benefit of KL regularization on replayed prompts.

\noindent\textit{Comparison of Selection Methods.} \textbf{Joint contrastive scoring improves replay selection.} As shown in Table~\ref{tab:ablation-score}, joint contrastive selection consistently outperforms accuracy-gain selection, selection based on either $P_1$ or $P_2$ alone. An intuitive alternative is to replay prompts on which the model has improved most, as measured by the gain in sampled accuracy. CoLiR's advantage over this alternative suggests that changes in the response distribution provide useful selection information beyond the gain in sampled accuracy, supporting the use of log-likelihood changes as a proxy for advantage. Joint scoring also outperforms scoring based on either $P_1$ or $P_2$ alone, indicating the benefit of considering how the model both reinforces correct responses and suppresses incorrect ones when selecting prompts for replay.

\subsection{Further Analysis}
\label{sec:further-analysis}
\textit{Likelihood Components of Selected Prompts.} \textbf{What likelihood changes characterize the prompts selected by CoLiR?} Selected prompts generally have high $P_1$ and $P_2$, indicating increased likelihood of correct responses and decreased likelihood of incorrect ones (Figure~\ref{fig:likelihood-components}). Meanwhile, the joint score also captures prompts with high $P_1$ but lower $P_2$, and vice versa. Thus, contrastive selection captures learning progress reflected in either or both components. These patterns are consistent with the contrastive score capturing a shift in relative preference toward correct responses, supporting the benefit of combining both components observed in Table~\ref{tab:ablation-score}.

\noindent\textit{Correctness Patterns of Selected Prompts.} \textbf{How does likelihood-based selection relate to changes in correctness?} As shown in Figure~\ref{fig:selection-rate}, selection rates vary across prompts grouped by early and late correctness, estimated from eight responses per prompt. On Navigation, selected prompts are concentrated in groups with high late correctness, while on Puzzle, selection favors prompts that move from low early correctness to higher late correctness and also show an accuracy increase. Selected prompts in We-Math2 and FinMME span a wider range of early-to-late correctness transitions. These observations suggest a correlation between likelihood changes and accuracy gains, while potentially capturing additional information about the response distribution beyond sampled accuracy.

}
\newcommand{\CoLiRReplayCapacityAppendix}{%
\section{Replay Memory Capacity}
\label{expapp-memory}
\textit{Replay Memory Capacity.} \textbf{How does replay memory capacity affect performance?} As shown in Table~\ref{tab:memory-capacity}, increasing the budget from 40 to 80 prompts per historical task improves acquisition and retention, whereas expanding it to 160 brings no further benefit. The strongest results at an intermediate capacity suggest that effective replay depends on prompt selection as well as memory capacity. This supports allocating a limited replay budget to prompts associated with substantial policy improvement.

\begin{center}
\begin{minipage}{.9\columnwidth}
\captionsetup{type=table}
\captionof{table}{\ExpMemoryCaption}
\label{tab:memory-capacity}
\begingroup
\small
\setlength{\tabcolsep}{3pt}
\renewcommand{\arraystretch}{1.10}
\makebox[\linewidth][c]{%
\begin{tabular}{>{\centering\arraybackslash}p{\dimexpr.40\linewidth-2\tabcolsep\relax}>{\centering\arraybackslash}p{\dimexpr.15\linewidth-2\tabcolsep\relax}>{\centering\arraybackslash}p{\dimexpr.15\linewidth-2\tabcolsep\relax}>{\centering\arraybackslash}p{\dimexpr.15\linewidth-2\tabcolsep\relax}}
\toprule
\textbf{Examples per task} & \textbf{MFN$\uparrow$} & \textbf{MTA$\uparrow$} & \textbf{MFT$\uparrow$} \\
\midrule
40 & 58.23 & 63.21 & 70.04 \\
80 & \textbf{69.35} & \textbf{70.86} & \textbf{72.27} \\
160 & 67.95 & 69.34 & 71.07 \\
\bottomrule
\end{tabular}%
}
\endgroup

\end{minipage}
\end{center}

}

\ifdefined\CoLiRPrepareOnly\else\CoLiRExperiments\fi

\let\CoLiRPrepareOnly\undefined

\maketitle
\begin{abstract}
Continual reinforcement learning (CRL) requires models to acquire new capabilities while retaining performance on previous tasks. A stability-oriented perspective common in conventional continual learning emphasizes knowledge preservation and interference mitigation. Complementing this perspective, we revisit CRL for vision-language models through policy improvement, a fundamental principle of RL, to preserve acquired performance gains. Using the performance difference lemma, we characterize historical performance changes through the current policy's expected advantages on historical task distributions, motivating replay of historical prompts with the current policy. To allocate limited replay memory, we propose Contrastive Likelihood Replay (CoLiR), which contrasts likelihood gains on correct responses from the final task policy with likelihood reductions on incorrect responses from the initial policy. Our analysis links correct-response likelihood gains to reference-policy advantages under an idealized KL-regularized optimal update and provides an endogenous-reward interpretation for general policy updates. CoLiR estimates scores from finite training rollouts and cached log-likelihoods, selects prompts at task boundaries, and replays them uniformly during subsequent training, with KL regularization restricted to replayed prompts. Memory construction only requires a small amount of additional cross-policy scoring but no additional response generation. Across three Qwen3-VL model sizes on MRCL, CoLiR achieves the highest mean final accuracy and mean task accuracy among the compared methods while maintaining competitive new-task performance.
\end{abstract}    

\begin{figure}[!t]
    \centering
    \includegraphics[width=\columnwidth]{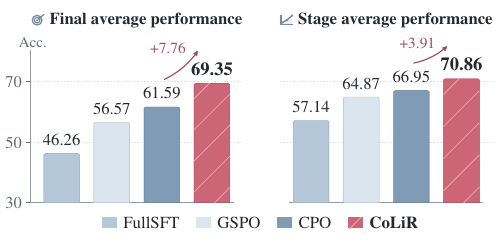}
    \caption{Task retention in CRL:
    CoLiR outperforms all compared methods in both final
    and stage-averaged accuracy.}
    \label{fig:motivation}
\end{figure}

\section{Introduction}
\label{sec:introduction}

Continual post-training enables large language and vision-language models (VLMs) to incrementally acquire diverse downstream capabilities in response to evolving application requirements~\cite{ctfsl, hide}.
However, adaptation to new tasks can degrade performance on previously
learned ones, a phenomenon known as catastrophic forgetting~\cite{ssr}. The central challenge lies in enabling cumulative capability acquisition across successive stages of post-training while preserving previously learned capabilities.

Prior efforts to mitigate forgetting in continual supervised fine-tuning have primarily relied on explicit knowledge-preservation mechanisms, such as rehearsal and parameter regularization~\cite{ssr,ewc}. 
Recent studies further suggest that the post-training paradigm itself influences knowledge retention~\cite{rft}.
Compared with supervised fine-tuning, reinforcement learning (RL) can exhibit stronger retention, attributed to on-policy data and mode-seeking updates~\cite{retaining}, as well as an implicit preference for solutions with smaller policy divergence~\cite{rlrazor}. 
These findings highlight the inherent potential of RL to mitigate forgetting, motivating further research into continual reinforcement learning (CRL) for large models.
Nevertheless, CPO reveals that RL remains susceptible to catastrophic forgetting in continual post-training of VLMs and introduces explicit regularization to constrain historical policy drift during sequential RL~\cite{cpo}.

These studies highlight the retention properties of RL and the effectiveness of constraining historical policy drift. 
However, such constraints do not directly characterize how policy updates affect historical task performance. This reflects a stability-oriented perspective commonly adopted in conventional continual learning, motivating a complementary perspective grounded in policy improvement, a fundamental principle of RL. Although RL pursues policy improvement by maximizing the expected return on an individual task, optimizing this objective independently at each learning stage does not guarantee the preservation of performance gains achieved on preceding tasks. This raises an important question: how can policy improvement guide the selection and retention of historical experiences during continual RL, particularly under limited memory capacity?

To address this question, we revisit CRL through the performance difference lemma, a foundational tool for analyzing policy improvement~\cite{trpo}.
Under a single-step formulation that treats a complete response as an action, the performance change on a historical task equals the current policy's expected advantage over a reference policy under that task's prompt distribution. 
Preserving previously acquired performance gains thus requires maintaining nonnegative expected advantages relative to the corresponding historical reference policies.
Current-task-only RL directly optimizes expected returns over the current task distribution, providing no guarantee of preserving performance on previous tasks.
Historical prompt replay addresses this limitation by allowing the current policy to respond to retained prompts and receive task-specific verifier feedback, thereby incorporating historical reward objectives into subsequent policy optimization.
Under limited memory capacity, however, the effectiveness of replay also depends on which historical prompts are retained.

From a policy improvement perspective, we aim to retain historical prompts associated with substantial performance gains during task learning, rather than relying solely on their final correctness or difficulty. 
However, directly estimating prompt-specific advantages requires evaluating the reference policy's expected reward, which can be computationally expensive.
To address this, we investigate policy likelihood changes as a tractable proxy for advantage-guided selection. 
Under an idealized KL-regularized update, we establish a monotonic relationship between likelihood gains on correct responses and their reference-policy advantages.
For general policy updates, we further characterize how likelihood gains reflect changes in success probability and probability redistribution among correct responses, providing an endogenous-reward interpretation~\cite{generalist_rm}. 
Together, these results provide theoretical motivation for using likelihood changes to guide historical prompt selection.

Building on this analysis, we propose \textbf{CoLiR} (\textbf{Contrastive Likelihood Replay}), a policy improvement-guided replay method for CRL.
CoLiR combines likelihood gains on correct responses sampled from the final task policy with likelihood reductions on incorrect responses sampled from the initial policy, capturing both the reinforcement of correct responses and the suppression of incorrect ones. 
We derive sufficient conditions under which the population contrastive score indicates positive policy improvement. In practice, we estimate this score from finite training rollouts and use the resulting ranking to construct a capacity-limited replay buffer at the end of each task.
During subsequent training, retained prompts are uniformly sampled for joint optimization with current-task data, with KL regularization applied to replayed prompts. 
By reusing training rollouts and cached log-likelihoods, CoLiR requires only a single round of additional cross-policy scoring, without generating new responses for memory construction.
Across three model sizes on MRCL, CoLiR achieves the highest mean final accuracy and mean task accuracy among the compared methods, while maintaining competitive performance on new tasks.
\section{Related Work}
\label{sec:related_work}

\paragraph{Continual Reinforcement Learning and Post-Training}
CRL aims to adapt policies to sequential tasks while preserving capabilities and enabling knowledge transfer~\cite{crlsurvey,crldefinition,continualworld}. 
Existing methods mitigated forgetting by constraining changes to important parameters (EWC~\cite{ewc}), consolidating knowledge through distillation (P\&C~\cite{pnc}), or combining replay with on-policy learning to balance retention and adaptation (CLEAR~\cite{er}). LAMOL and SSR explored generated training examples for knowledge retention in language-model continual learning~\cite{lamol,ssr}.
In continual post-training, prior studies found that RL could retain knowledge better than supervised fine-tuning~\cite{rft,retaining}, which RL’s Razor attributed to a preference for smaller policy divergence~\cite{rlrazor}.
Yet CPO showed that RL can still suffer substantial forgetting in multimodal reasoning and proposed replay-free regularization to constrain historical policy drift~\cite{cpo}.
While prior work emphasized knowledge retention and interference mitigation, we revisit CRL for VLMs through the performance difference lemma and develop likelihood-based selective replay to retain previously acquired policy improvements.

\paragraph{Replay Selection}
In conventional RL, PER assigned higher sampling probabilities to transitions with larger temporal-difference errors~\cite{per}, while PLR prioritized environments by estimated learning potential~\cite{plr}.
For efficient language-model RL, Prompt Replay prioritized intermediate-difficulty prompts for repeated on-policy optimization~\cite{promptreplay}; Prompt Curriculum Learning and RIF-RFT selected prompts with effective learning signals~\cite{pcl,rft}; and DPS and GPS used historical training dynamics to predict informative prompts and reduce sampling overhead~\cite{dps,gps}. 
CPR retains historical prompts and dynamically prioritized their reuse by recent pass rates and cooldown constraints~\cite{cpr}. 
These prompt-selection methods primarily targeted difficulty or future learning potential. 
To retain prior policy improvements in CRL for VLMs, we draw on connections between policy likelihood ratios and implicit rewards~\cite{dpo,generalist_rm} to select historical prompts with pronounced likelihood changes during task learning. 
Unlike dynamic replay prioritization, our memory selection occurs at task boundaries and is decoupled from subsequent uniform replay.
\section{Preliminaries}
\label{sec:preliminary}
\paragraph{Reinforcement Learning with Verifiable Rewards}
RLVR optimizes language models with programmatically checkable outcomes ~\cite{grpo, dapo}. For a prompt \(x\sim D\), the policy generates \(y=(o,a)\), comprising a reasoning trace \(o\) and final answer \(a\); a task-specific verifier assigns the binary correctness reward \(R(x,y)\in\{0,1\}\). The rollout policy \(\pi_{\mathrm{old}}\equiv\pi_{\theta_{\mathrm{old}}}\) samples \(G\) responses \(Y=\{y^j\}_{j=1}^{G}\) per prompt. Group Relative Policy Optimization (GRPO)~\cite{grpo} estimates group-relative advantages:
\begin{equation*}
\widehat A_j
=
\frac{R(x,y^j)-\mu_R}{\sigma_R+\xi},
\end{equation*}
where \(\mu_R\) and \(\sigma_R\) are the group reward mean and standard deviation, and \(\xi>0\) ensures numerical stability. 
All tokens in a response share this advantage under outcome supervision. Define the clipped surrogate

\begin{equation*}
C_\epsilon(r,A)
=
\min\left\{
rA,\operatorname{clip}(r,1-\epsilon,1+\epsilon)A
\right\},
\end{equation*}
with clipping parameter \(\epsilon\). 
GRPO applies token-level clipping, whereas CoLiR uses Group Sequence Policy Optimization (GSPO)~\cite{gspo} as the optimization algorithm, which applies sequence-level clipping for training stability, with length-normalized importance ratios and objective
\begin{equation*}
J(\pi_\theta)
=
\frac{1}{G}\mathbb E_{x,Y}
\left[
\sum_{j=1}^{G}
C_\epsilon(\left[
\frac{\pi_\theta(y^j\mid x)}
{\pi_{\mathrm{old}}(y^j\mid x)}
\right]^{1/|y^j|},\widehat A_j)
\right].
\end{equation*}

\paragraph{Continual Reinforcement Learning}
CRL trains a shared policy on sequential tasks \(\mathcal T_1,\ldots,\mathcal T_K\). 
Each VLM task \(i\) provides prompts \(x=(v,u)\sim D_i\), with visual input \(v\) and textual query \(u\). 
From an RL perspective, we model response generation as a single-step decision: \(x\) is the initial state, and the complete response \(y\sim\pi_\theta(\cdot\mid x)\) is the action, scored by a task verifier with \(R_i(x,y)\in\{0,1\}\). 
Let \(p_i^\pi(x)=
\mathbb E_{y\sim\pi(\cdot\mid x)}[R_i(x,y)]\)
denote the prompt-level success probability.
The expected task return is
\begin{equation*}
    J_i(\pi_\theta)
=
\mathbb E_{x\sim D_i,\;y\sim\pi_\theta(\cdot\mid x)}
[R_i(x,y)],
\end{equation*}
which equals expected answer accuracy under binary rewards. 
At stage \(k\), the goal is to improve performance on \(D_k\) while mitigating forgetting on tasks \(i<k\). 
Historical performance is \(J_{\mathrm{old}}(\pi_\theta)=\sum_{i<k}w_iJ_i(\pi_\theta)\), with fixed evaluation weights \(w_i\ge0\) and \(\sum_{i<k}w_i=1\).
We use a streaming setting: training data arrive task by task, and direct access to historical data is restricted after each task.

\section{Methodology}
\label{sec:methodology}

\subsection{Revisiting CRL through Policy Improvement}

We revisit CRL through the lens of policy improvement, a fundamental principle of RL.
For a historical task $i$ and a fixed reference policy $\bar\pi_i$, the response-level advantage is
\begin{equation}
A_i^{\bar\pi_i}(x,y)
=
R_i(x,y)-p_i^{\bar\pi_i}(x).
\label{eq:advantage}
\end{equation}
Under the single-step formulation, the performance difference lemma~\cite{trpo} yields
\begin{equation}
J_i(\pi_\theta)-J_i(\bar\pi_i)
=
\mathbb E_{\substack{
x\sim\mathcal D_i,
y\sim\pi_\theta(\cdot\mid x)}}
\left[A_i^{\bar\pi_i}(x,y)\right].
\label{eq:pdl}
\end{equation}

\noindent\textbf{Cross-task perspective.}
Eq.~\eqref{eq:pdl} indicates preserving or improving historical performance requires nonnegative expected advantage relative to the reference policy, evaluated over historical inputs $x\sim\mathcal D_i$ and responses generated by the current policy $y\sim\pi_\theta(\cdot\mid x)$.
Since the reference baseline $p_i^{\bar\pi_i}(x)$ is independent of $\theta$, maximizing the expected advantage is equivalent to maximizing the expected reward on task $i$.
Optimizing only the current distribution $\mathcal D_k$, which generally differs from $\mathcal D_i$, does not guarantee historical performance. This motivates historical task replay, which reintroduces past task inputs and enables the current policy to be optimized against historical task objectives.

\noindent\textbf{Within-task perspective.}
Limited memory capacity of historical replay raises a further question: \textit{which historical prompts should be retained?}
From a policy improvement perspective, advantage measures the relative value of a response, and differences in response advantages reflect the varying contributions of prompts.
This motivates prioritizing prompts that contributed most to advantage maximization, with the aim of preserving the key improvements acquired during previous tasks.
However, binary verification rewards indicate correctness without directly quantifying the advantage relative to the reference policy, making such contributions difficult to assess directly.
This motivates exploring more accessible signals that can reflect advantage-related learning contributions and guide replay selection.

\subsection{Policy Likelihood Ratios as an Advantage Proxy}
To identify historical prompts associated with substantial policy improvement, we investigate whether policy likelihood changes of correct responses can serve as a tractable proxy for advantage-based replay selection.

Let $\pi_i^0$ and $\pi_i^T$ denote the policies before and after learning task $i$, respectively.
For each prompt $x$, let $q_T^+(\cdot\mid x)$ denote the final policy's conditional distribution over correct responses.
We consider the average log-likelihood improvement of correct responses:
\begin{equation}
P_1(x)
=
\mathbb E_{y\sim q_T^+(\cdot\mid x)}
\left[
\log\frac{\pi_i^T(y\mid x)}
{\pi_i^0(y\mid x)}
\right].
\label{eq:p1-definition}
\end{equation}
We establish its theoretical motivation through advantage monotonicity and endogenous reward interpretation.

\paragraph{Connection to Policy Advantage.}
We first consider an idealized KL-regularized policy improvement objective:
\begin{equation}
\pi_i^T
=
\arg\max_\pi
\left\{
\mathbb E_{y\sim\pi}[R_i(x,y)]
-
\beta D_{\mathrm{KL}}(\pi\Vert\pi_i^0)
\right\},
\end{equation}
where $\beta>0$ is a fixed regularization coefficient.

\begin{proposition}[Advantage Monotonicity]
\label{prop:AM}
Under the KL-regularized optimal policy, $P_1(x)$ is strictly increasing in the reference-policy advantage of a correct response:
\begin{equation}
P_1(x)
=
-\log\left[
1-A_i^{\pi_i^0}(x,y^+)
(1-e^{-1/\beta})
\right].
\label{eq:p1-advantage}
\end{equation}
\end{proposition}

\noindent The proof is provided in Appendix A.1.
For sufficiently strong KL regularization, $P_1(x)\approx A_i^{\pi_i^0}(x,y^+)/\beta$.
This idealized result motivates likelihood-based selection, although GSPO does not require KL regularization.
Proposition~\ref{prop:ERPD} provides a complementary endogenous-reward interpretation for general policy updates.

\paragraph{Endogenous Reward Interpretation.}
We next provide an endogenous-reward interpretation of $P_1$ that applies to general policy updates.
Recent work on endogenous rewards~\cite{generalist_rm} connects language model likelihoods to implicit rewards through inverse reinforcement learning.
Following its outcome-reward formulation, we consider
\begin{equation}
\widehat r_\pi(x,y)
=
\alpha\log\pi(y\mid x)+V_\pi(x),
\label{eq:endogenous-reward}
\end{equation}
where $\alpha>0$ and $V_\pi(x)$ is a prompt-dependent value term.

\begin{proposition}[Endogenous Reward and Performance Decomposition]
\label{prop:ERPD}
Let $p_t(x)$ and $q_t^+(\cdot\mid x)$ denote
the success probability and conditional
distribution over correct responses under
$\pi_i^t$, for $t\in\{0,T\}$.
Assuming $0<p_0(x),p_T(x)<1$ and common support
on correct responses, we have
\begin{equation}
P_1(x)
=
\frac{
\mathbb E_{q_T^+}[\Delta\widehat r_i(x,y)]
-\Delta V_i(x)
}{\alpha},
\label{eq:p1-endogenous}
\end{equation}
and
\begin{equation}
P_1(x)
=
\log\frac{p_T(x)}{p_0(x)}
+
D_{\mathrm{KL}}(q_T^+\Vert q_0^+).
\label{eq:p1-performance}
\end{equation}
\end{proposition}

\noindent The proof is provided in Appendix A.2.
The first identity interprets $P_1$ as the
response-dependent component of endogenous
reward changes.
The second decomposes $P_1$ into the
log-relative change in success probability
and the distributional shift among correct
responses.
Under the KL-regularized update in Proposition~\ref{prop:AM}, the conditional distribution over correct responses is unchanged, making the KL term vanish. 
For general policy updates, this term captures additional changes in the distribution of correct responses.

Combining Eq.~\eqref{eq:p1-performance} with the performance difference lemma gives \begin{equation}
\mathbb E_{\pi_i^T}
[A_i^{\pi_i^0}(x,y)]
=
p_0(x)
\left[
e^{P_1(x)-D_{\mathrm{KL}}(q_T^+\Vert q_0^+)}
-1
\right].
\label{eq:p1-pdl-connection}
\end{equation}
Thus, when the KL divergence term is small, $P_1$ closely approximates the log-relative performance improvement.

Together, these results justify $P_1$ as a tractable proxy for advantage-guided replay selection. Under the KL-regularized policy improvement, $P_1$ preserves the ordering of reference-policy advantages.
Under general policy updates, it captures the response-dependent component of endogenous reward changes and is explicitly related to the realized expected advantage through Eq.~\eqref{eq:p1-pdl-connection}.

\begin{figure*}[ht]
\centering
\includegraphics[width=\textwidth]{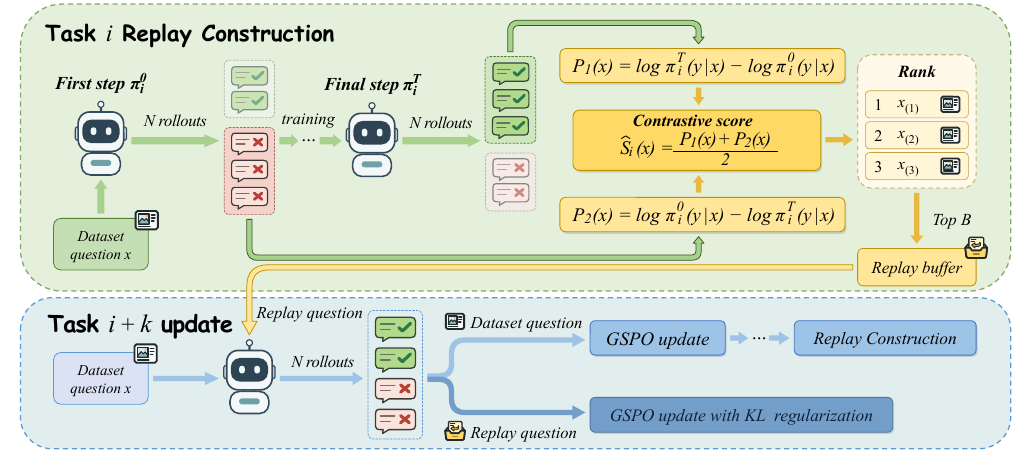}
\caption{Overview of CoLiR. CoLiR builds a replay buffer based on contrastive scores and samples from it during training on new tasks.}
\label{fig:method}
\end{figure*}

\subsection{CoLiR: Contrastive Likelihood Replay}
While the preceding analysis establishes the theoretical motivation for $P_1$, policy improvement under binary rewards involves both reinforcing correct responses and suppressing incorrect ones.
Specifically, relative to the initial policy $\pi_i^0$, incorrect responses have negative advantages, i.e., $A_i^{\pi_i^0}(x,y^-)=-p_i^{\pi_i^0}(x)$.
To capture this complementary aspect, we define
\begin{equation}
P_2(x)
=
\mathbb E_{y\sim q_0^-(\cdot\mid x)}
\left[
\log\frac{\pi_i^0(y\mid x)}
{\pi_i^T(y\mid x)}
\right],
\label{eq:p2-definition}
\end{equation}
where $q_0^-$ denotes the initial policy's conditional distribution over incorrect responses.
Thus, $P_1$ measures the reinforcement of correct responses generated by the final policy, whereas $P_2$ measures the suppression of incorrect responses generated by the initial policy.

We combine these complementary signals into a contrastive score:
\begin{equation}
S_i(x)=\frac{1}{2}\left(P_1(x)+P_2(x)\right).
\label{eq:replay-score}
\end{equation}
Following Eq.~\eqref{eq:endogenous-reward}, we have
\begin{equation}
2\alpha S_i(x)=\mathbb E_{q_T^+}[\Delta\widehat r_i(x,y)]-\mathbb E_{q_0^-}[\Delta\widehat r_i(x,y)].
\label{eq:contrastive-endogenous}
\end{equation}
This contrast cancels the prompt-dependent value change, allowing $S_i(x)$ to capture shifts in endogenous reward preference for correct over incorrect responses.
Furthermore, the contrastive score admits a direct connection to realized policy improvement.
\CoLiRMainResultsTable
\CoLiROODTable
\begin{proposition}[Contrastive Likelihood and Policy Improvement]
\label{prop:CLPI}
Let $p_t(x)$ and $q_t^\pm(\cdot\mid x)$ denote
the success probability and conditional response
distributions under $\pi_i^t$, for $t\in\{0,T\}$.
Assume $0<p_0(x),p_T(x)<1$ and finite
conditional KL divergences.
Then
\begin{equation}
2S_i(x)
=
\operatorname{logit}p_T(x)
-\operatorname{logit}p_0(x)
+\varepsilon_i(x),
\label{eq:contrastive-decomposition}
\end{equation}
where
\begin{equation*}
\varepsilon_i(x)
=
D_{\mathrm{KL}}(q_T^+\Vert q_0^+)
+
D_{\mathrm{KL}}(q_0^-\Vert q_T^-)
\geq 0.
\end{equation*}
If $\varepsilon_i(x)\leq\delta$, then
\begin{equation}
S_i(x)>\frac{\delta}{2}
\Longrightarrow
\mathbb E_{y\sim\pi_i^T(\cdot\mid x)}
[A_i^{\pi_i^0}(x,y)]>0.
\label{eq:contrastive-advantage}
\end{equation}
\end{proposition}
\noindent The proof is provided in Appendix A.3.
Proposition~\ref{prop:CLPI} relates the contrastive score to realized policy improvement through changes in success log-odds.
Under bounded distributional discrepancies, larger scores provide higher lower bounds on policy improvement, motivating the selection of high-scoring prompts for replay.

In practical implementation, we leverage the group-based sampling mechanism, which generates $G$ responses per prompt and records their log-likelihoods during policy optimization.
To reduce computational cost, for each task $i$, we reuse the incorrect responses from the initial training iteration and the correct responses from the final iteration, together with their corresponding log-likelihoods.
Let $\mathcal Y_{0,x}^-$ and $\mathcal Y_{T,x}^+$ denote these two response sets, respectively.
The empirical contrastive score is computed as
\begin{equation}
\begin{aligned}
\widehat S_i(x)
=&\frac{1}{2|\mathcal Y_{T,x}^+|}
\sum_{y\in\mathcal Y_{T,x}^+}
\log\frac{\pi_i^T(y\mid x)}{\pi_i^0(y\mid x)}\\
&+\frac{1}{2|\mathcal Y_{0,x}^-|}
\sum_{y\in\mathcal Y_{0,x}^-}
\log\frac{\pi_i^0(y\mid x)}{\pi_i^T(y\mid x)}.
\end{aligned}
\label{eq:empirical-replay-score}
\end{equation}
Since the log-likelihoods under the generating policies are already available, only two additional cross-policy likelihood evaluations on partial responses are required, without additional response generation.

After completing task $i$, we consider prompts whose final rollout groups have mixed correctness outcomes, and whose initial rollout groups contain at least one incorrect response.
We rank these eligible prompts by $\widehat S_i(x)$ and retain the top-$B$ prompts in the replay buffer, where $B$ denotes the per-task memory budget.
During subsequent task learning, we allocate a replay budget and randomly sample historical prompts from the replay buffer for joint optimization with current-task data. To align with the KL-regularized policy improvement, we further impose a KL term on replayed prompts to mitigate forgetting.
\CoLiRExperiments
\section{Conclusions}
\label{sec:conclusion}
We revisit CRL for VLMs through policy improvement, using the performance difference lemma to motivate historical prompt replay. CoLiR selects prompts by contrasting likelihood gains on correct final-policy responses with reductions on incorrect initial-policy responses. Our analysis connects correct-response likelihood gains to reference-policy advantages under idealized KL-regularized optimization and provides an endogenous-reward interpretation for general updates. CoLiR estimates scores from cached training rollouts and likelihoods, requiring no additional response generation for selection. The selected prompts are uniformly replayed during subsequent RL training. Experiments across three model scales show improved retention while maintaining new-task learning.
\section*{LLM Usage Statement}
We used OpenAI’s GPT-6 and GPT-6.1 to assist with text polishing, related-work searches and citation checks, experimental analysis, and checking theorem proofs. For figures, these models only drafted initial layouts, which the authors manually refined. The authors take full responsibility for all submitted content.
We provided major portions of the manuscript to OpenAI for reviewer-style feedback. 
The authors take full responsibility for all submitted content.

{
    \small
    \bibliographystyle{ieeenat_fullname}
    \bibliography{main}

@String(ICLR = {Int. Conf. Learn. Represent.})

@String(ICLR  = {ICLR})

@article{grpo,
  author       = {Zhihong Shao and
                  Peiyi Wang and
                  Qihao Zhu and
                  Runxin Xu and
                  Junxiao Song and
                  Mingchuan Zhang and
                  Y. K. Li and
                  Y. Wu and
                  Daya Guo},
  title        = {DeepSeekMath: Pushing the Limits of Mathematical Reasoning in Open
                  Language Models},
  journal      = {CoRR},
  volume       = {abs/2402.03300},
  year         = {2024},
  url          = {https://doi.org/10.48550/arXiv.2402.03300},
  doi          = {10.48550/ARXIV.2402.03300},
  eprinttype   = {arXiv},
  eprint       = {2402.03300},
  bibsource    = {dblp computer science bibliography, https://dblp.org}
}

@inproceedings{dapo,
  author       = {Qiying Yu and
                  Zheng Zhang and
                  Ruofei Zhu and
                  Yufeng Yuan and
                  Xiaochen Zuo and
                  Yu Yue and
                  Weinan Dai and
                  Tiantian Fan and
                  Gaohong Liu and
                  Juncai Liu and
                  Lingjun Liu and
                  Xin Liu and
                  Haibin Lin and
                  Zhiqi Lin and
                  Bole Ma and
                  Guangming Sheng and
                  Yuxuan Tong and
                  Chi Zhang and
                  Mofan Zhang and
                  Ru Zhang and
                  Wang Zhang and
                  Hang Zhu and
                  Jinhua Zhu and
                  Jiaze Chen and
                  Jiangjie Chen and
                  Chengyi Wang and
                  Hongli Yu and
                  Yuxuan Song and
                  Xiangpeng Wei and
                  Hao Zhou and
                  Jingjing Liu and
                  Wei{-}Ying Ma and
                  Ya{-}Qin Zhang and
                  Lin Yan and
                  Yonghui Wu and
                  Mingxuan Wang},
  editor       = {Danielle Belgrave and
                  Cheng Zhang and
                  Laura N. Montoya and
                  Hsuan{-}Tien Lin and
                  Razvan Pascanu and
                  Piotr Koniusz and
                  Marzyeh Ghassemi and
                  Nancy Chen and
                  Iv{\'{a}}n Vladimir Meza Ru{\'{\i}}z and
                  Arturo Loaiza{-}Bonilla},
  title        = {{DAPO:} An Open-Source {LLM} Reinforcement Learning System at Scale},
  booktitle    = {Advances in Neural Information Processing Systems 38: Annual Conference
                  on Neural Information Processing Systems 2025, NeurIPS 2025, San Diego,
                  CA, USA, December 2-7, 2025 / Mexico City, Mexico, November 30 - December
                  5, 2025},
  year         = {2025},
  url          = {http://papers.nips.cc/paper\_files/paper/2025/hash/a4277440d50f1f15d2cb4c14f7e0c0d2-Abstract-Conference.html},
  bibsource    = {dblp computer science bibliography, https://dblp.org}
}

@article{gspo,
  author       = {Chujie Zheng and
                  Shixuan Liu and
                  Mingze Li and
                  Xiong{-}Hui Chen and
                  Bowen Yu and
                  Chang Gao and
                  Kai Dang and
                  Yuqiong Liu and
                  Rui Men and
                  An Yang and
                  Jingren Zhou and
                  Junyang Lin},
  title        = {Group Sequence Policy Optimization},
  journal      = {CoRR},
  volume       = {abs/2507.18071},
  year         = {2025},
  url          = {https://doi.org/10.48550/arXiv.2507.18071},
  doi          = {10.48550/ARXIV.2507.18071},
  eprinttype   = {arXiv},
  eprint       = {2507.18071},
  bibsource    = {dblp computer science bibliography, https://dblp.org}
}

@article{cpo,
  author       = {Mao{-}Lin Luo and
                  Zhe{-}Xu Wang and
                  Zi{-}Hao Zhou and
                  Bo Ye and
                  Jian Zhao and
                  Min{-}Ling Zhang and
                  Tong Wei},
  title        = {{RL} Forgets! Towards Continual Policy Optimization},
  journal      = {CoRR},
  volume       = {abs/2607.04364},
  year         = {2026},
  url          = {https://doi.org/10.48550/arXiv.2607.04364},
  doi          = {10.48550/ARXIV.2607.04364},
  eprinttype   = {arXiv},
  eprint       = {2607.04364},
  bibsource    = {dblp computer science bibliography, https://dblp.org}
}

@inproceedings{trpo,
  author       = {John Schulman and
                  Sergey Levine and
                  Pieter Abbeel and
                  Michael I. Jordan and
                  Philipp Moritz},
  editor       = {Francis R. Bach and
                  David M. Blei},
  title        = {Trust Region Policy Optimization},
  booktitle    = {Proceedings of the 32nd International Conference on Machine Learning,
                  {ICML} 2015, Lille, France, 6-11 July 2015},
  series       = {{JMLR} Workshop and Conference Proceedings},
  volume       = {37},
  pages        = {1889--1897},
  publisher    = {JMLR.org},
  year         = {2015},
  url          = {http://proceedings.mlr.press/v37/schulman15.html},
  bibsource    = {dblp computer science bibliography, https://dblp.org}
}

@article{generalist_rm,
  author       = {Yi{-}Chen Li and
                  Tian Xu and
                  Yang Yu and
                  Xuqin Zhang and
                  Xiong{-}Hui Chen and
                  Zhongxiang Ling and
                  Ningjing Chao and
                  Lei Yuan and
                  Zhi{-}Hua Zhou},
  title        = {Generalist Reward Models: Found Inside Large Language Models},
  journal      = {CoRR},
  volume       = {abs/2506.23235},
  year         = {2025},
  url          = {https://doi.org/10.48550/arXiv.2506.23235},
  doi          = {10.48550/ARXIV.2506.23235},
  eprinttype   = {arXiv},
  eprint       = {2506.23235},
  bibsource    = {dblp computer science bibliography, https://dblp.org}
}

@article{crlsurvey,
  author       = {Khimya Khetarpal and
                  Matthew Riemer and
                  Irina Rish and
                  Doina Precup},
  title        = {Towards Continual Reinforcement Learning: {A} Review and Perspectives},
  journal      = {J. Artif. Intell. Res.},
  volume       = {75},
  pages        = {1401--1476},
  year         = {2022},
  url          = {https://doi.org/10.1613/jair.1.13673},
  doi          = {10.1613/JAIR.1.13673},
  bibsource    = {dblp computer science bibliography, https://dblp.org}
}

@inproceedings{crldefinition,
  author       = {David Abel and
                  Andr{\'{e}} Barreto and
                  Benjamin Van Roy and
                  Doina Precup and
                  Hado Philip van Hasselt and
                  Satinder Singh},
  editor       = {Alice Oh and
                  Tristan Naumann and
                  Amir Globerson and
                  Kate Saenko and
                  Moritz Hardt and
                  Sergey Levine},
  title        = {A Definition of Continual Reinforcement Learning},
  booktitle    = {Advances in Neural Information Processing Systems 36: Annual Conference
                  on Neural Information Processing Systems 2023, NeurIPS 2023, New Orleans,
                  LA, USA, December 10 - 16, 2023},
  year         = {2023},
  url          = {http://papers.nips.cc/paper\_files/paper/2023/hash/9d8cf1247786d6dfeefeeb53b8b5f6d7-Abstract-Conference.html},
  bibsource    = {dblp computer science bibliography, https://dblp.org}
}

@inproceedings{continualworld,
  author       = {Maciej Wolczyk and
                  Michal Zajac and
                  Razvan Pascanu and
                  Lukasz Kucinski and
                  Piotr Milos},
  editor       = {Marc'Aurelio Ranzato and
                  Alina Beygelzimer and
                  Yann N. Dauphin and
                  Percy Liang and
                  Jennifer Wortman Vaughan},
  title        = {Continual World: {A} Robotic Benchmark For Continual Reinforcement
                  Learning},
  booktitle    = {Advances in Neural Information Processing Systems 34: Annual Conference
                  on Neural Information Processing Systems 2021, NeurIPS 2021, December
                  6-14, 2021, virtual},
  pages        = {28496--28510},
  year         = {2021},
  url          = {https://proceedings.neurips.cc/paper/2021/hash/ef8446f35513a8d6aa2308357a268a7e-Abstract.html},
  bibsource    = {dblp computer science bibliography, https://dblp.org}
}

@article{ewc,
  author       = {Brandon Shuen Yi Loke and
                  Filippo Quadri and
                  Gabriel Vivanco and
                  Maximilian Casagrande and
                  Sa{\'{u}}l Fenollosa},
  title        = {Overcoming catastrophic forgetting in neural networks},
  journal      = {CoRR},
  volume       = {abs/2507.10485},
  year         = {2025},
  url          = {https://doi.org/10.48550/arXiv.2507.10485},
  doi          = {10.48550/ARXIV.2507.10485},
  eprinttype   = {arXiv},
  eprint       = {2507.10485},
  bibsource    = {dblp computer science bibliography, https://dblp.org}
}

@inproceedings{pnc,
  author       = {Jonathan Schwarz and
                  Wojciech Czarnecki and
                  Jelena Luketina and
                  Agnieszka Grabska{-}Barwinska and
                  Yee Whye Teh and
                  Razvan Pascanu and
                  Raia Hadsell},
  editor       = {Jennifer G. Dy and
                  Andreas Krause},
  title        = {Progress {\&} Compress: {A} scalable framework for continual learning},
  booktitle    = {Proceedings of the 35th International Conference on Machine Learning,
                  {ICML} 2018, Stockholmsm{\"{a}}ssan, Stockholm, Sweden, July
                  10-15, 2018},
  series       = {Proceedings of Machine Learning Research},
  volume       = {80},
  pages        = {4535--4544},
  publisher    = {{PMLR}},
  year         = {2018},
  url          = {http://proceedings.mlr.press/v80/schwarz18a.html},
  bibsource    = {dblp computer science bibliography, https://dblp.org}
}

@inproceedings{er,
  author       = {David Rolnick and
                  Arun Ahuja and
                  Jonathan Schwarz and
                  Timothy P. Lillicrap and
                  Gregory Wayne},
  editor       = {Hanna M. Wallach and
                  Hugo Larochelle and
                  Alina Beygelzimer and
                  Florence d'Alch{\'{e}}{-}Buc and
                  Emily B. Fox and
                  Roman Garnett},
  title        = {Experience Replay for Continual Learning},
  booktitle    = {Advances in Neural Information Processing Systems 32: Annual Conference
                  on Neural Information Processing Systems 2019, NeurIPS 2019, December
                  8-14, 2019, Vancouver, BC, Canada},
  pages        = {348--358},
  year         = {2019},
  url          = {https://proceedings.neurips.cc/paper/2019/hash/fa7cdfad1a5aaf8370ebeda47a1ff1c3-Abstract.html},
  bibsource    = {dblp computer science bibliography, https://dblp.org}
}

@inproceedings{lamol,
  author       = {Fan{-}Keng Sun and
                  Cheng{-}Hao Ho and
                  Hung{-}Yi Lee},
  title        = {{LAMOL:} LAnguage MOdeling for Lifelong Language Learning},
  booktitle    = {8th International Conference on Learning Representations, {ICLR} 2020,
                  Addis Ababa, Ethiopia, April 26-30, 2020},
  publisher    = {OpenReview.net},
  year         = {2020},
  url          = {https://openreview.net/forum?id=Skgxcn4YDS},
  bibsource    = {dblp computer science bibliography, https://dblp.org}
}

@inproceedings{ssr,
  author       = {Jianheng Huang and
                  Leyang Cui and
                  Ante Wang and
                  Chengyi Yang and
                  Xinting Liao and
                  Linfeng Song and
                  Junfeng Yao and
                  Jinsong Su},
  editor       = {Lun{-}Wei Ku and
                  Andre Martins and
                  Vivek Srikumar},
  title        = {Mitigating Catastrophic Forgetting in Large Language Models with Self-Synthesized
                  Rehearsal},
  booktitle    = {Proceedings of the 62nd Annual Meeting of the Association for Computational
                  Linguistics (Volume 1: Long Papers), {ACL} 2024, Bangkok, Thailand,
                  August 11-16, 2024},
  pages        = {1416--1428},
  publisher    = {Association for Computational Linguistics},
  year         = {2024},
  url          = {https://doi.org/10.18653/v1/2024.acl-long.77},
  doi          = {10.18653/V1/2024.ACL-LONG.77},
  bibsource    = {dblp computer science bibliography, https://dblp.org}
}

@article{rft,
  author       = {Song Lai and
                  Haohan Zhao and
                  Rong Feng and
                  Changyi Ma and
                  Wenzhuo Liu and
                  Hongbo Zhao and
                  Xi Lin and
                  Dong Yi and
                  Min Xie and
                  Qingfu Zhang and
                  Hongbin Liu and
                  Gaofeng Meng and
                  Fei Zhu},
  title        = {Reinforcement Fine-Tuning Naturally Mitigates Forgetting in Continual
                  Post-Training},
  journal      = {CoRR},
  volume       = {abs/2507.05386},
  year         = {2025},
  url          = {https://doi.org/10.48550/arXiv.2507.05386},
  doi          = {10.48550/ARXIV.2507.05386},
  eprinttype   = {arXiv},
  eprint       = {2507.05386},
  bibsource    = {dblp computer science bibliography, https://dblp.org}
}

@article{retaining,
  author       = {Howard Chen and
                  Noam Razin and
                  Karthik Narasimhan and
                  Danqi Chen},
  title        = {Retaining by Doing: The Role of On-Policy Data in Mitigating Forgetting},
  journal      = {CoRR},
  volume       = {abs/2510.18874},
  year         = {2025},
  url          = {https://doi.org/10.48550/arXiv.2510.18874},
  doi          = {10.48550/ARXIV.2510.18874},
  eprinttype   = {arXiv},
  eprint       = {2510.18874},
  bibsource    = {dblp computer science bibliography, https://dblp.org}
}

@inproceedings{rlrazor,
  author       = {Idan Shenfeld and
                  Jyothish Pari and
                  Pulkit Agrawal},
  editor       = {Carl Vondrick and
                  Bharath Hariharan and
                  Colin Raffel and
                  Lerrel Pinto and
                  Diyi Yang and
                  Aleksandra Faust},
  title        = {RL's Razor: Why Online Reinforcement Learning Forgets Less},
  booktitle    = {The Fourteenth International Conference on Learning Representations,
                  {ICLR} 2026, Rio de Janeiro, Brazil, April 23-27, 2026},
  publisher    = {proceedings.iclr.cc},
  year         = {2026},
  url          = {https://proceedings.iclr.cc/paper\_files/paper/2026/hash/618c95f4557c15b253fb0e6f548ea0c0-Abstract-Conference.html},
  bibsource    = {dblp computer science bibliography, https://dblp.org}
}

@inproceedings{per,
  author       = {Tom Schaul and
                  John Quan and
                  Ioannis Antonoglou and
                  David Silver},
  editor       = {Yoshua Bengio and
                  Yann LeCun},
  title        = {Prioritized Experience Replay},
  booktitle    = {4th International Conference on Learning Representations, {ICLR} 2016,
                  San Juan, Puerto Rico, May 2-4, 2016, Conference Track Proceedings},
  year         = {2016},
  url          = {https://doi.org/10.48550/arXiv.1511.05952},
  doi          = {10.48550/ARXIV.1511.05952},
  bibsource    = {dblp computer science bibliography, https://dblp.org}
}

@inproceedings{plr,
  author       = {Minqi Jiang and
                  Edward Grefenstette and
                  Tim Rockt{\"{a}}schel},
  editor       = {Marina Meila and
                  Tong Zhang},
  title        = {Prioritized Level Replay},
  booktitle    = {Proceedings of the 38th International Conference on Machine Learning,
                  {ICML} 2021, 18-24 July 2021, Virtual Event},
  series       = {Proceedings of Machine Learning Research},
  volume       = {139},
  pages        = {4940--4950},
  publisher    = {{PMLR}},
  year         = {2021},
  url          = {http://proceedings.mlr.press/v139/jiang21b.html},
  bibsource    = {dblp computer science bibliography, https://dblp.org}
}

@article{promptreplay,
  author       = {Andrei Baroian and
                  Rutger Berger},
  title        = {Prompt replay: speeding up grpo with on-policy reuse of high-signal
                  prompts},
  journal      = {CoRR},
  volume       = {abs/2603.21177},
  year         = {2026},
  url          = {https://doi.org/10.48550/arXiv.2603.21177},
  doi          = {10.48550/ARXIV.2603.21177},
  eprinttype   = {arXiv},
  eprint       = {2603.21177},
  bibsource    = {dblp computer science bibliography, https://dblp.org}
}

@inproceedings{pcl,
  author       = {Zhaolin Gao and
                  Joongwon Kim and
                  Wen Sun and
                  Thorsten Joachims and
                  Sid Wang and
                  Richard Yuanzhe Pang and
                  Liang Tan},
  editor       = {Carl Vondrick and
                  Bharath Hariharan and
                  Colin Raffel and
                  Lerrel Pinto and
                  Diyi Yang and
                  Aleksandra Faust},
  title        = {Prompt Curriculum Learning for Efficient {LLM} Post-Training},
  booktitle    = {The Fourteenth International Conference on Learning Representations,
                  {ICLR} 2026, Rio de Janeiro, Brazil, April 23-27, 2026},
  publisher    = {proceedings.iclr.cc},
  year         = {2026},
  url          = {https://proceedings.iclr.cc/paper\_files/paper/2026/hash/601b42cebb3650ddb075770e2a85f428-Abstract-Conference.html},
  bibsource    = {dblp computer science bibliography, https://dblp.org}
}

@inproceedings{dps,
  author       = {Yixiu Mao and
                  Yun Qu and
                  Qi Wang and
                  Heming Zou and
                  Xiangyang Ji},
  editor       = {Carl Vondrick and
                  Bharath Hariharan and
                  Colin Raffel and
                  Lerrel Pinto and
                  Diyi Yang and
                  Aleksandra Faust},
  title        = {Dynamics-Predictive Sampling for Active {RL} Finetuning of Large Reasoning
                  Models},
  booktitle    = {The Fourteenth International Conference on Learning Representations,
                  {ICLR} 2026, Rio de Janeiro, Brazil, April 23-27, 2026},
  publisher    = {proceedings.iclr.cc},
  year         = {2026},
  url          = {https://proceedings.iclr.cc/paper\_files/paper/2026/hash/2ae33575c3374050654ae7802326c81d-Abstract-Conference.html},
  bibsource    = {dblp computer science bibliography, https://dblp.org}
}

@article{gps,
  author       = {Yun Qu and
                  Qi Wang and
                  Yixiu Mao and
                  Heming Zou and
                  Yuhang Jiang and
                  Weijie Liu and
                  Clive Bai and
                  Kai Yang and
                  Yangkun Chen and
                  Saiyong Yang and
                  Xiangyang Ji},
  title        = {Small Generalizable Prompt Predictive Models Can Steer Efficient {RL}
                  Post-Training of Large Reasoning Models},
  journal      = {CoRR},
  volume       = {abs/2602.01970},
  year         = {2026},
  url          = {https://doi.org/10.48550/arXiv.2602.01970},
  doi          = {10.48550/ARXIV.2602.01970},
  eprinttype   = {arXiv},
  eprint       = {2602.01970},
  bibsource    = {dblp computer science bibliography, https://dblp.org}
}

@article{cpr,
  author       = {Lirui Luo and
                  Guoxi Zhang and
                  Hongming Xu and
                  Rongqing Li and
                  Cong Fang and
                  Lifeng Fan},
  title        = {Continual Reasoning Gym: Diagnosing and Harnessing Shared Reasoning
                  in Continual {RLVR}},
  journal      = {CoRR},
  volume       = {abs/2608.18574},
  year         = {2026},
  url          = {https://doi.org/10.48550/arXiv.2608.18574},
  doi          = {10.48550/ARXIV.2608.18574},
  eprinttype   = {arXiv},
  eprint       = {2608.18574},
  bibsource    = {dblp computer science bibliography, https://dblp.org}
}

@inproceedings{dpo,
  author       = {Rafael Rafailov and
                  Archit Sharma and
                  Eric Mitchell and
                  Christopher D. Manning and
                  Stefano Ermon and
                  Chelsea Finn},
  editor       = {Alice Oh and
                  Tristan Naumann and
                  Amir Globerson and
                  Kate Saenko and
                  Moritz Hardt and
                  Sergey Levine},
  title        = {Direct Preference Optimization: Your Language Model is Secretly a
                  Reward Model},
  booktitle    = {Advances in Neural Information Processing Systems 36: Annual Conference
                  on Neural Information Processing Systems 2023, NeurIPS 2023, New Orleans,
                  LA, USA, December 10 - 16, 2023},
  year         = {2023},
  url          = {http://papers.nips.cc/paper\_files/paper/2023/hash/a85b405ed65c6477a4fe8302b5e06ce7-Abstract-Conference.html},
  bibsource    = {dblp computer science bibliography, https://dblp.org}
}

@inproceedings{hide,
  author       = {Haiyang Guo and
                  Fanhu Zeng and
                  Ziwei Xiang and
                  Fei Zhu and
                  Da{-}Han Wang and
                  Xu{-}Yao Zhang and
                  Cheng{-}Lin Liu},
  editor       = {Wanxiang Che and
                  Joyce Nabende and
                  Ekaterina Shutova and
                  Mohammad Taher Pilehvar},
  title        = {HiDe-LLaVA: Hierarchical Decoupling for Continual Instruction Tuning
                  of Multimodal Large Language Model},
  booktitle    = {Proceedings of the 63rd Annual Meeting of the Association for Computational
                  Linguistics (Volume 1: Long Papers), {ACL} 2025, Vienna, Austria,
                  July 27 - August 1, 2025},
  pages        = {13572--13586},
  publisher    = {Association for Computational Linguistics},
  year         = {2025},
  url          = {https://doi.org/10.18653/v1/2025.acl-long.666},
  doi          = {10.18653/V1/2025.ACL-LONG.666},
  bibsource    = {dblp computer science bibliography, https://dblp.org}
}

@inproceedings{ctfsl,
  author       = {Zixuan Ke and
                  Haowei Lin and
                  Yijia Shao and
                  Hu Xu and
                  Lei Shu and
                  Bing Liu},
  editor       = {Yoav Goldberg and
                  Zornitsa Kozareva and
                  Yue Zhang},
  title        = {Continual Training of Language Models for Few-Shot Learning},
  booktitle    = {Proceedings of the 2022 Conference on Empirical Methods in Natural
                  Language Processing, {EMNLP} 2022, Abu Dhabi, United Arab Emirates,
                  December 7-11, 2022},
  pages        = {10205--10216},
  publisher    = {Association for Computational Linguistics},
  year         = {2022},
  url          = {https://doi.org/10.18653/v1/2022.emnlp-main.695},
  doi          = {10.18653/V1/2022.EMNLP-MAIN.695},
  bibsource    = {dblp computer science bibliography, https://dblp.org}
}

@inproceedings{lora,
  author       = {Edward J. Hu and
                  Yelong Shen and
                  Phillip Wallis and
                  Zeyuan Allen{-}Zhu and
                  Yuanzhi Li and
                  Shean Wang and
                  Lu Wang and
                  Weizhu Chen},
  title        = {LoRA: Low-Rank Adaptation of Large Language Models},
  booktitle    = {The Tenth International Conference on Learning Representations, {ICLR}
                  2022, Virtual Event, April 25-29, 2022},
  publisher    = {OpenReview.net},
  year         = {2022},
  url          = {https://openreview.net/forum?id=nZeVKeeFYf9},
  bibsource    = {dblp computer science bibliography, https://dblp.org}
}

@inproceedings{olora,
  author       = {Xiao Wang and
                  Tianze Chen and
                  Qiming Ge and
                  Han Xia and
                  Rong Bao and
                  Rui Zheng and
                  Qi Zhang and
                  Tao Gui and
                  Xuanjing Huang},
  editor       = {Houda Bouamor and
                  Juan Pino and
                  Kalika Bali},
  title        = {Orthogonal Subspace Learning for Language Model Continual Learning},
  booktitle    = {Findings of the Association for Computational Linguistics: {EMNLP}
                  2023, Singapore, December 6-10, 2023},
  series       = {Findings of {ACL}},
  volume       = {{EMNLP} 2023},
  pages        = {10658--10671},
  publisher    = {Association for Computational Linguistics},
  year         = {2023},
  url          = {https://doi.org/10.18653/v1/2023.findings-emnlp.715},
  doi          = {10.18653/V1/2023.FINDINGS-EMNLP.715},
  bibsource    = {dblp computer science bibliography, https://dblp.org}
}

@inproceedings{reglora,
  author       = {Jinpeng Chen and
                  Runmin Cong and
                  Yuzhi Zhao and
                  Hongzheng Yang and
                  Guangneng Hu and
                  Horace H. S. Ip and
                  Sam Kwong},
  editor       = {Aarti Singh and
                  Maryam Fazel and
                  Daniel Hsu and
                  Simon Lacoste{-}Julien and
                  Felix Berkenkamp and
                  Tegan Maharaj and
                  Kiri Wagstaff and
                  Jerry Zhu},
  title        = {{SEFE:} Superficial and Essential Forgetting Eliminator for Multimodal
                  Continual Instruction Tuning},
  booktitle    = {Forty-second International Conference on Machine Learning, {ICML}
                  2025, Vancouver, BC, Canada, July 13-19, 2025},
  series       = {Proceedings of Machine Learning Research},
  volume       = {267},
  publisher    = {{PMLR} / OpenReview.net},
  year         = {2025},
  url          = {https://proceedings.mlr.press/v267/chen25n.html},
  bibsource    = {dblp computer science bibliography, https://dblp.org}
}

@article{mmmupro,
 title={{MMMU-Pro: A More Robust Multi-discipline Multimodal Understanding Benchmark}},
 author={Xiang Yue and Tianyu Zheng and Yuansheng Ni and Yubo Wang and Kai Zhang and Shengbang Tong and Yuxuan Sun and Botao Yu and Ge Zhang and Huan Sun and Yu Su and Wenhu Chen and Graham Neubig},
 journal={arXiv preprint arXiv:2409.02813},
 year={2024},
 url={https://arxiv.org/abs/2409.02813}
}

@article{mathvista,
 title={{MathVista: Evaluating Mathematical Reasoning of Foundation Models in Visual Contexts}},
 author={Pan Lu and Hritik Bansal and Tony Xia and Jiacheng Liu and Chunyuan Li and Hannaneh Hajishirzi and Hao Cheng and Kai-Wei Chang and Michel Galley and Jianfeng Gao},
 journal={arXiv preprint arXiv:2310.02255},
 year={2023},
 url={https://arxiv.org/abs/2310.02255}
}

@article{mathvision,
 title={{Measuring Multimodal Mathematical Reasoning with MATH-Vision Dataset}},
 author={Ke Wang and Junting Pan and Weikang Shi and Zimu Lu and Mingjie Zhan and Hongsheng Li},
 journal={arXiv preprint arXiv:2402.14804},
 year={2024},
 url={https://arxiv.org/abs/2402.14804}
}

@article{mathverse,
 title={{MathVerse: Does Your Multi-modal LLM Truly See the Diagrams in Visual Math Problems?}},
 author={Renrui Zhang and Dongzhi Jiang and Yichi Zhang and Haokun Lin and Ziyu Guo and Pengshuo Qiu and Aojun Zhou and Pan Lu and Kai-Wei Chang and Peng Gao and Hongsheng Li},
 journal={arXiv preprint arXiv:2403.14624},
 year={2024},
 url={https://arxiv.org/abs/2403.14624}
}

@article{mmstar,
 title={{Are We on the Right Way for Evaluating Large Vision-Language Models?}},
 author={Lin Chen and Jinsong Li and Xiaoyi Dong and Pan Zhang and Yuhang Zang and Zehui Chen and Haodong Duan and Jiaqi Wang and Yu Qiao and Dahua Lin and Feng Zhao},
 journal={arXiv preprint arXiv:2403.20330},
 year={2024},
 url={https://arxiv.org/abs/2403.20330}
}

@article{charxiv,
 title={{CharXiv: Charting Gaps in Realistic Chart Understanding in Multimodal LLMs}},
 author={Zirui Wang and Mengzhou Xia and Luxi He and Howard Chen and Yitao Liu and Richard Zhu and Kaiqu Liang and Xindi Wu and Haotian Liu and Sadhika Malladi and Alexis Chevalier and Sanjeev Arora and Danqi Chen},
 journal={arXiv preprint arXiv:2406.18521},
 year={2024},
 url={https://arxiv.org/abs/2406.18521}
}

@article{countbench,
 title={{Teaching CLIP to Count to Ten}},
 author={Roni Paiss and Ariel Ephrat and Omer Tov and Shiran Zada and Inbar Mosseri and Michal Irani and Tali Dekel},
 journal={arXiv preprint arXiv:2302.12066},
 year={2023},
 url={https://arxiv.org/abs/2302.12066}
}

@article{paligemma,
 title={{PaliGemma: A versatile 3B VLM for transfer}},
 author={Lucas Beyer and Andreas Steiner and Andr{\'e} Susano Pinto and Alexander Kolesnikov and Xiao Wang and Daniel Salz and Maxim Neumann and Ibrahim Alabdulmohsin and Michael Tschannen and Emanuele Bugliarello and Thomas Unterthiner and Daniel Keysers and Skanda Koppula and Fangyu Liu and Adam Grycner and Alexey Gritsenko and Neil Houlsby and Manoj Kumar and Keran Rong and Julian Eisenschlos and Rishabh Kabra and Matthias Bauer and Matko Bo{\v s}njak and Xi Chen and Matthias Minderer and Paul Voigtlaender and Ioana Bica and Ivana Balazevic and Joan Puigcerver and Pinelopi Papalampidi and Olivier Henaff and Xi Xiong and Radu Soricut and Jeremiah Harmsen and Xiaohua Zhai},
 journal={arXiv preprint arXiv:2407.07726},
 year={2024},
 url={https://arxiv.org/abs/2407.07726}
}

@inproceedings{pope,
 title={Evaluating Object Hallucination in Large Vision-Language Models},
 author={Yifan Li and Yifan Du and Kun Zhou and Jinpeng Wang and Wayne Xin Zhao and Ji-Rong Wen},
 booktitle={Proceedings of the 2023 Conference on Empirical Methods in Natural Language Processing},
 year={2023},
 url={https://aclanthology.org/2023.emnlp-main.20/}
}

@inproceedings{docvqa,
 title={{DocVQA}: A Dataset for {VQA} on Document Images},
 author={Minesh Mathew and Dimosthenis Karatzas and C. V. Jawahar},
 booktitle={Proceedings of the IEEE/CVF Winter Conference on Applications of Computer Vision},
 year={2021},
 pages={2200--2209},
 url={https://openaccess.thecvf.com/content/WACV2021/html/Mathew_DocVQA_A_Dataset_for_VQA_on_Document_Images_WACV_2021_paper.html}
}

@misc{realworldqa,
 title={{Grok-1.5 Vision Preview}},
 author={{xAI}},
 year={2024},
 howpublished={\url{https://x.ai/news/grok-1.5v}}
}
}

\appendix
\clearpage
\setcounter{page}{1}
\maketitlesupplementary

\section{Theoretical Proofs}
\subsection{Proof of Proposition~\ref{prop:AM}}
\label{proof:AM}
\begin{proof}
    The optimal policy associated with the KL-regularized objective has the
Boltzmann form
\begin{equation*}
\pi_i^T(y\mid x)
=
\frac{
\pi_i^0(y\mid x)e^{R_i(x,y)/\beta}
}{
Z_i(x)
},
\end{equation*}
where the normalization term is
\begin{equation*}
Z_i(x)
=
1-p_i^{\pi_i^0}(x)
+p_i^{\pi_i^0}(x)e^{1/\beta}.
\end{equation*}
For any correct response $y^+$ under the binary reward formulation,
$R_i(x,y^+)=1$. Its advantage under the reference policy is therefore
\begin{equation*}
A_i^{\pi_i^0}(x,y^+)
=
R_i(x,y^+)
-
\mathbb E_{y\sim\pi_i^0(\cdot\mid x)}[R_i(x,y)]
=
1-p_i^{\pi_i^0}(x).
\end{equation*}
Moreover, substituting the optimal-policy expression into
Eq.~\eqref{eq:p1-definition} gives
\begin{equation*}
P_1(x)
=
\log
\frac{
e^{1/\beta}
}{
1-p_i^{\pi_i^0}(x)
+p_i^{\pi_i^0}(x)e^{1/\beta}
}.
\end{equation*}
Using
$p_i^{\pi_i^0}(x)=1-A_i^{\pi_i^0}(x,y^+)$, we obtain
\begin{equation*}
P_1(x)
=
-\log\left[
1-
A_i^{\pi_i^0}(x,y^+)
\left(1-e^{-1/\beta}\right)
\right],
\end{equation*}
which establishes Eq.~\eqref{eq:p1-advantage}. 

Furthermore, differentiating this expression with respect to the reference-policy advantage yields
\begin{equation*}
\frac{\partial P_1(x)}
{\partial A_i^{\pi_i^0}(x,y^+)}
=
\frac{1-e^{-1/\beta}}
{1-A_i^{\pi_i^0}(x,y^+)
(1-e^{-1/\beta})}
>0.
\end{equation*}
Hence, $P_1(x)$ is strictly increasing in the positive-response advantage.

When $\beta$ is sufficiently large, $1/\beta$ is small. Applying the
first-order Taylor expansion $e^{-z}=1-z+O(z^2)$ around $z=0$ gives
\begin{equation*}
1-e^{-1/\beta}
=
\frac{1}{\beta}
+
O\left(\frac{1}{\beta^2}\right)
\approx
\frac{1}{\beta}.
\end{equation*}
Substituting this approximation into Eq.~\eqref{eq:p1-advantage} and applying
$-\log(1-z)=z+O(z^2)$ for small $z$ yields
\begin{equation*}
P_1(x)
\approx
\frac{
A_i^{\pi_i^0}(x,y^+)
}{\beta}.
\end{equation*}
Thus, under strong KL regularization, the likelihood improvement of correct
responses is approximately proportional to the reference-policy advantage.
\end{proof}

\subsection{Proof of Proposition~\ref{prop:ERPD}}
\label{proof:ERPD}
\begin{proof}
    We first prove the endogenous reward characterization. By the definition
of the endogenous reward,
\[
\widehat r_{\pi_i^T}(x,y)
=
\alpha\log\pi_i^T(y\mid x)+V_{\pi_i^T}(x),
\]
and
\[
\widehat r_{\pi_i^0}(x,y)
=
\alpha\log\pi_i^0(y\mid x)+V_{\pi_i^0}(x).
\]
Taking their difference gives
\begin{equation*}
\Delta\widehat r_i(x,y)
=
\alpha
\log\frac{\pi_i^T(y\mid x)}
{\pi_i^0(y\mid x)}
+
\Delta V_i(x).
\end{equation*}
Since $\Delta V_i(x)$ depends only on the prompt and is independent of the
response $y$, taking the expectation under the final policy's correct-response
distribution $q_T^+(\cdot\mid x)$ yields
\begin{equation*}
\mathbb E_{q_T^+}
\left[
\Delta\widehat r_i(x,y)
\right]
=
\alpha P_1(x)+\Delta V_i(x).
\end{equation*}
Rearranging this equality proves Eq.~\eqref{eq:p1-endogenous}.

We next derive the performance decomposition. For any correct response
$y\in\mathcal Y_x^+$, the policy probability can be factorized as
\[
\pi_i^t(y\mid x)
=
p_t(x)\,q_t^+(y\mid x),
\qquad t\in\{0,T\},
\]
because $p_t(x)$ is the total probability assigned to correct responses and
$q_t^+(\cdot\mid x)$ is the corresponding conditional distribution. Therefore,
\begin{equation*}
\frac{\pi_i^T(y\mid x)}
{\pi_i^0(y\mid x)}
=
\frac{p_T(x)}{p_0(x)}
\cdot
\frac{q_T^+(y\mid x)}
{q_0^+(y\mid x)}.
\end{equation*}
Taking logarithms gives
\begin{equation*}
\log\frac{\pi_i^T(y\mid x)}
{\pi_i^0(y\mid x)}
=
\log\frac{p_T(x)}{p_0(x)}
+
\log\frac{q_T^+(y\mid x)}
{q_0^+(y\mid x)}.
\end{equation*}
Taking the expectation with respect to
$q_T^+(\cdot\mid x)$, the first term remains unchanged because it is independent
of $y$, while the second term becomes
\begin{equation*}
\mathbb E_{q_T^+}
\left[
\log\frac{q_T^+(y\mid x)}
{q_0^+(y\mid x)}
\right]
=
D_{\mathrm{KL}}
\left(
q_T^+(\cdot\mid x)
\Vert
q_0^+(\cdot\mid x)
\right).
\end{equation*}
Combining the two terms proves Eq.~\eqref{eq:p1-performance}.
\end{proof}

\subsection{Proof of Proposition~\ref{prop:CLPI}}
\label{proof:CLPI}

\begin{proof}
For a given prompt $x$, let
$\mathcal Y_x^+$ and $\mathcal Y_x^-$ denote
the sets of correct and incorrect responses,
respectively.
By the definition of conditional distributions,
for $t\in\{0,T\}$, we have
\begin{equation}
\pi_i^t(y\mid x)=
\begin{cases}
p_t(x)q_t^+(y\mid x), & y\in\mathcal Y_x^+,\\
(1-p_t(x))q_t^-(y\mid x), & y\in\mathcal Y_x^-.
\end{cases}
\end{equation}

For correct responses sampled from the final policy,
the likelihood improvement can be expressed as
\begin{equation}
\begin{aligned}
P_1(x)
&=
\mathbb E_{q_T^+}
\left[
\log\frac{\pi_i^T(y\mid x)}
{\pi_i^0(y\mid x)}
\right]\\
&=
\log\frac{p_T(x)}{p_0(x)}
+
\mathbb E_{q_T^+}
\left[
\log\frac{q_T^+(y\mid x)}
{q_0^+(y\mid x)}
\right]\\
&=
\log\frac{p_T(x)}{p_0(x)}
+
D_{\mathrm{KL}}(q_T^+\Vert q_0^+).
\end{aligned}
\end{equation}
Similarly, for incorrect responses sampled from
the initial policy, we obtain
\begin{equation}
\begin{aligned}
P_2(x)
&=
\mathbb E_{q_0^-}
\left[
\log\frac{\pi_i^0(y\mid x)}
{\pi_i^T(y\mid x)}
\right]\\
&=
\log\frac{1-p_0(x)}{1-p_T(x)}
+
D_{\mathrm{KL}}(q_0^-\Vert q_T^-).
\end{aligned}
\end{equation}

Combining the two expressions yields
\begin{equation}
\begin{aligned}
2S_i(x)
&=P_1(x)+P_2(x)\\
&=\log\frac{p_T(x)(1-p_0(x))}
{p_0(x)(1-p_T(x))}
+\varepsilon_i(x)\\
&=\operatorname{logit}p_T(x)
-\operatorname{logit}p_0(x)
+\varepsilon_i(x),
\end{aligned}
\end{equation}
where
\begin{equation*}
\varepsilon_i(x)
=
D_{\mathrm{KL}}(q_T^+\Vert q_0^+)
+
D_{\mathrm{KL}}(q_0^-\Vert q_T^-)
\geq0.
\end{equation*}
This establishes
Eq.~\eqref{eq:contrastive-decomposition}.

Since $\varepsilon_i(x)$ is nonnegative,
the decomposition immediately implies
\begin{equation}
\begin{aligned}
p_T(x)>p_0(x)&\Longrightarrow S_i(x)>0,\\
S_i(x)<0&\Longrightarrow p_T(x)<p_0(x).
\end{aligned}
\end{equation}
Thus, positive policy improvement necessarily
yields a positive score, whereas a negative
score indicates performance degradation.

Furthermore, suppose
$\varepsilon_i(x)\leq\delta$.
Rearranging the decomposition gives
\begin{equation}
\begin{aligned}
\operatorname{logit}p_T(x)
-\operatorname{logit}p_0(x)
&=2S_i(x)-\varepsilon_i(x)\\
&\geq2S_i(x)-\delta.
\end{aligned}
\end{equation}
Therefore, if $S_i(x)>\delta/2$, then
$\operatorname{logit}p_T(x)>
\operatorname{logit}p_0(x)$.
Since the logit function is strictly increasing,
this implies $p_T(x)>p_0(x)$.

Finally, by the performance difference lemma
under the single-step formulation,
\begin{equation}
\begin{aligned}
&\mathbb E_{y\sim\pi_i^T(\cdot\mid x)}
[A_i^{\pi_i^0}(x,y)]\\
&=
\mathbb E_{y\sim\pi_i^T(\cdot\mid x)}
[R_i(x,y)-p_0(x)]\\
&=p_T(x)-p_0(x)>0.
\end{aligned}
\end{equation}
Hence, under bounded distributional discrepancy,
a sufficiently large contrastive score guarantees
a positive expected advantage, completing the proof.
\end{proof}

\begingroup
\setlength{\intextsep}{4pt plus 1pt minus 1pt}
\setlength{\floatsep}{6pt plus 1pt minus 1pt}
\setlength{\textfloatsep}{6pt plus 1pt minus 1pt}
\section{Experimental Details}
\label{app:experimental-details}

\subsection{Datasets and Corrections}
Figure~\ref{fig:mrcl-task-examples} shows one test example from each MRCL task in training order, illustrating the different response formats used across tasks. For MedBookVQA, an extra space in correct options leaks the answers, allowing all evaluated models to easily exceed 90\% accuracy. We standardize option spacing to remove this shortcut. For Navigation, we correct an answer annotation and extend scoring to accept equivalent shortest paths. For We-Math2, we repair malformed text and incorrect answers, remove duplicate records, and exclude invalid examples. For Puzzle, we remove duplicate examples. For FinMME, we correct answer annotations, remove empty unit hints and duplicate records, and exclude invalid examples. The scoring implementation supports accepted-answer aliases so that equivalent valid answers receive the same reward. Table~\ref{app:repair-counts} summarizes the four-task correction patch; answer corrections are counted before duplicate removal.

\begin{table}[h]
\centering\small
\caption{Summary of the data correction patch for T2--T5.}
\label{app:repair-counts}
\setlength{\tabcolsep}{4pt}
\begin{tabular}{lrrr}
\toprule
Task & Corrected & Excluded & Duplicates \\
\midrule
Navigation & 1 & 0 & 0 \\
We-Math2 & 19 & 13 & 138 \\
Puzzle & 0 & 0 & 2 \\
FinMME & 14 & 58 & 4 \\
\bottomrule
\end{tabular}
\end{table}

\subsection{Baseline Methods}
\label{app:baselines}
\paragraph{Supervised baselines.}
(1) \textbf{FullSFT}: updates all model parameters through supervised fine-tuning; (2) \textbf{LoRA}~\cite{lora}: learns low-rank weight updates while freezing the pretrained parameters; (3) \textbf{O-LoRA}~\cite{olora}: uses orthogonal low-rank subspaces to reduce interference between tasks; (4) \textbf{RegLoRA}~\cite{reglora}: regularizes important elements of LoRA weight updates to preserve historical knowledge.

\paragraph{Reinforcement learning baselines.}
(1) \textbf{GRPO}~\cite{grpo}: optimizes the policy using group-relative advantages; (2) \textbf{GRPO}$^{*}$~\cite{grpo}: adds KL regularization to GRPO; (3) \textbf{GSPO}~\cite{gspo}: uses sequence-level importance ratios for policy optimization; (4) \textbf{CPO}~\cite{cpo}: regularizes changes to important parameters to mitigate forgetting during sequential RL training.

\subsection{Implementation Details}
We train the supervised baselines for 2, 3, 1, 1, and 1 epochs on T1--T5, respectively, while RL training runs for 600 steps per task.
For all model sizes, GSPO, CPO, and CoLiR share the same checkpoint after training on MedBookVQA with GSPO, since the first stage involves only a single task. Their continual-learning updates begin at the second stage.

Table~\ref{app:colir-settings} summarizes the training and replay settings of CoLiR. The early and late response sets used for selection come from the first and last distinct complete generation events recorded for each prompt during task training. When fewer than $K$ prompts satisfy the filtering criteria, all eligible prompts are retained.

\begin{table}[h]
\centering\normalsize
\caption{Training and replay settings of CoLiR.}
\label{app:colir-settings}
\setlength{\tabcolsep}{5pt}
\renewcommand{\arraystretch}{1.05}
\newsavebox{\AppendixSettingsBox}
\sbox{\AppendixSettingsBox}{%
\begin{tabular}{>{\raggedright\arraybackslash}p{\dimexpr.72\linewidth-2\tabcolsep\relax}>{\centering\arraybackslash}p{\dimexpr.28\linewidth-2\tabcolsep\relax}}
\toprule
\textbf{Parameter Name} & \textbf{Value} \\
\midrule
Learning rate & $5\times10^{-6}$ \\
Training steps (per task) & 600 \\
Training batch size & 32 \\
Group size $G$ & 8 \\
Replay batch size (T2) & 4 \\
Replay batch size (T3--T5) & 8 \\
Replay buffer size $K$ (per task) & 80 \\
Replay KL coefficient $\beta$ & 0.01 \\
Clipping parameter $\epsilon_{\mathrm{low}}$ & $3\times10^{-4}$ \\
Clipping parameter $\epsilon_{\mathrm{high}}$ & $4\times10^{-4}$ \\
\bottomrule
\end{tabular}%
}
\begin{tikzpicture}
\node[inner sep=0pt,outer sep=0pt] (settings) {\usebox{\AppendixSettingsBox}};
\draw[line width=.4pt]
($(settings.south west)+(.72\linewidth,3pt)$) --
($(settings.north west)+(.72\linewidth,-22.5pt)$);
\draw[line width=.4pt]
($(settings.north west)+(.72\linewidth,-3pt)$) --
($(settings.north west)+(.72\linewidth,-15.5pt)$);
\end{tikzpicture}
\end{table}

\subsection{Evaluation Metrics}
Let $a_{t,i}$ denote accuracy on task $i$ after training on task $t$, with $1\leq i\leq t\leq T$ and $T=5$. The metrics introduced in the main text are computed as
\begin{align}
\mathrm{MFN} &= \frac{1}{T}\sum_{i=1}^{T}a_{T,i}, \\
\mathrm{MTA} &= \frac{1}{T}\sum_{i=1}^{T}\left(\frac{1}{T-i+1}\sum_{t=i}^{T}a_{t,i}\right), \\
\mathrm{MFT} &= \frac{1}{T}\sum_{i=1}^{T}a_{i,i}.
\end{align}

\section{Additional Experiments}
\label{app:additional-experiments}
\subsection{Replay Memory Capacity}
\label{expapp-memory}
\textbf{How does the number of retained prompts affect performance?} As shown in Table~\ref{tab:memory-capacity}, increasing the capacity from 40 to 80 prompts per historical task improves MFN, MTA, and MFT, with a larger gain in MFN than in MFT. Increasing the capacity further to 160 does not yield additional gains, and $K=80$ achieves the strongest results in this comparison. This pattern indicates that replay benefits do not increase monotonically with memory capacity. In CoLiR, capacity determines how many prompts are retained from the contrastive ranking, and expanding the retained set also includes lower-ranked candidates. The results are consistent with prioritizing prompts associated with substantial policy improvement, rather than relying solely on a larger replay memory.

\begin{table}[h]
\centering\small
\caption{Performance under different replay memory capacities. $K$ is the maximum number of retained prompts per historical task. Best results are shown in \textbf{bold}.}
\label{tab:memory-capacity}
\setlength{\tabcolsep}{9pt}
\renewcommand{\arraystretch}{1.1}
\begin{tabular}{crrr}
\toprule
$K$ & MFN$\uparrow$ & MTA$\uparrow$ & MFT$\uparrow$ \\
\midrule
40 & 58.23 & 63.21 & 70.04 \\
\rowcolor{CoLiRTableTint}80 & \textbf{69.35} & \textbf{70.86} & \textbf{72.27} \\
160 & 67.95 & 69.34 & 71.07 \\
\bottomrule
\end{tabular}
\end{table}

\subsection{Contrastive Score Ablation}
\label{app:checkpoint-branches}
\textbf{Do higher-scoring prompts provide greater replay value?}
We initialize three branches from the same 4B checkpoint saved at the end of Navigation training (T2), so all three inherit the same MedBookVQA replay set.
We rank the Navigation candidate prompts by contrastive score. One branch adds the top 80 prompts to its replay buffer, another adds the bottom 80, and the third adds 80 sampled at random from the same pool.
All three branches then train on We-Math2 (T3) under otherwise identical settings.

As shown in Table~\ref{tab:checkpoint-branches}, the three branches achieve similar accuracy on We-Math2, while high-score replay yields the highest accuracy on both prior tasks, with the clearest advantage on Navigation.
On Navigation, low-score replay also outperforms random replay.
Figure~4 in the main paper suggests that higher contrastive scores tend to accompany larger gains in prompt-level accuracy during the original task, while lower-scoring prompts tend to show smaller gains.
The low-score set may therefore contain more prompts for which accuracy remains relatively low late in Navigation training, including some with success rates near 50\%.
At this level, sampling several responses to the same prompt is more likely to yield both correct and incorrect answers within a rollout group.
The resulting differences in reward provide a learning signal during replay and may explain the low-score branch's advantage over random selection.
The further gain from high-score replay, together with the comparison of selection methods in Table~4 of the main paper, suggests that selecting prompts with high contrastive scores helps direct a limited replay budget toward prompts on which the policy made larger gains during the original task.

\begin{table}[!ht]
\centering\small
\caption{Task accuracy (\%) after T3 with different replay selections from a shared checkpoint. Best results are shown in \textbf{bold}.}
\label{tab:checkpoint-branches}
\setlength{\tabcolsep}{4pt}
\renewcommand{\arraystretch}{1.1}
\begin{tabular}{lccc}
\toprule
\multirow{2}{*}[-1.5pt]{Selection} & \multicolumn{2}{c}{Historical tasks} & Current task \\
\cmidrule(lr){2-3}\cmidrule(l){4-4}
& MedBookVQA & Navigation & We-Math2 \\
\midrule
\rowcolor{CoLiRTableTint}High-score & \textbf{69.27} & \textbf{91.28} & \textbf{63.26} \\
Low-score & 68.59 & 86.83 & 62.17 \\
Random & 64.86 & 84.60 & 62.53 \\
\bottomrule
\end{tabular}
\end{table}

\subsection{Composition of Selected Prompts}
\label{app:knowledge-composition}
\textbf{How does contrastive selection change the composition of the replay set?} As shown in Figure~\ref{fig:replay-composition}, the selected prompts differ from the eligible pool in both knowledge category and path length. On We-Math2, geometry accounts for a larger share of the selected set, increasing from 63.9\% to 79.2\%, while the share of algebra decreases and that of probability and statistics remains similar. Selection therefore changes the relative emphasis across knowledge categories while retaining prompts from all three categories. On Navigation, the selected set is more concentrated on paths of 2--4 steps, especially 3-step paths, whereas paths of 6 steps or more become less frequent. Thus, contrastive selection does not simply favor longer paths. These shifts arise after candidate filtering, when eligible prompts are ranked by their contrastive scores. Under our policy-improvement formulation, likelihood changes serve as a proxy for advantage; selection therefore prioritizes the improvement reflected in response distributions rather than preserving category proportions or ranking prompts by path length. The resulting replay set reflects where the policy has improved within each task, connecting the observed composition shifts to the selection criterion.

\par\smallskip
\noindent\begin{minipage}{\linewidth}
\centering
\includegraphics[width=.76\linewidth]{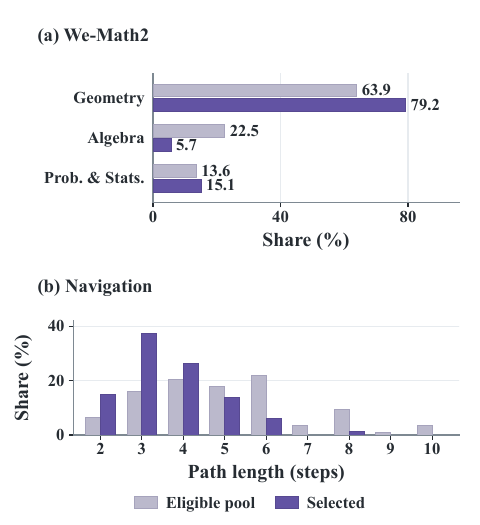}
\captionof{figure}{Composition of the eligible pool and selected replay prompts: (a) knowledge categories on We-Math2; (b) path lengths on Navigation. Shares are normalized within each pool.}
\label{fig:replay-composition}
\end{minipage}\par

\begin{figure}[p]
\centering
\includegraphics[width=.98\columnwidth]{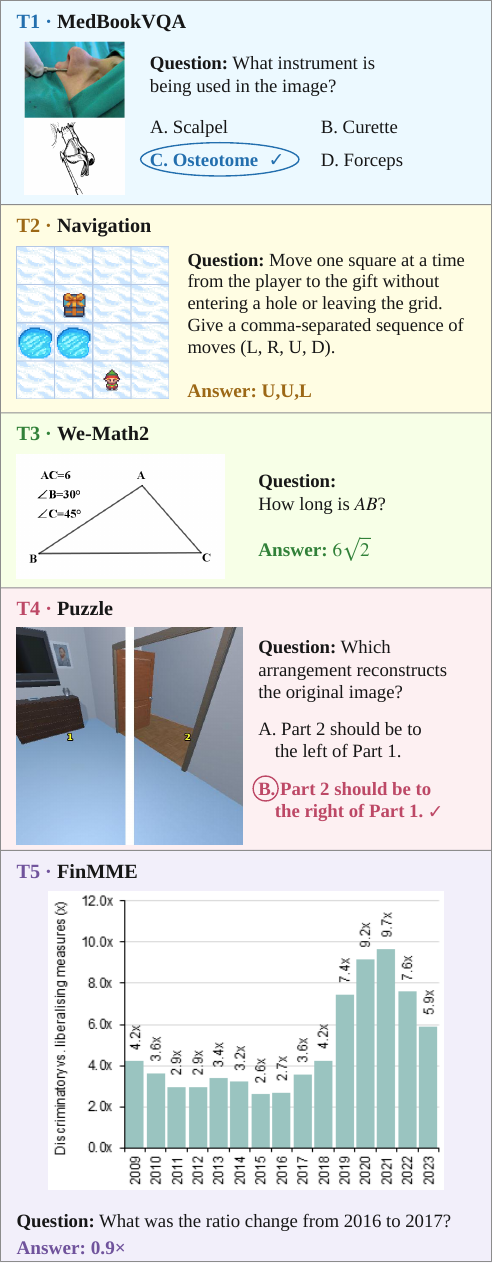}
\caption{Examples from the five MRCL~\cite{cpo} tasks in training order. Each panel shows a test image, its question, and the reference answer. Task instructions are abbreviated for display.}
\label{fig:mrcl-task-examples}
\end{figure}

\endgroup
\onecolumn
\section{Detailed Continual Learning Results}
\label{expapp-continual}
Tables~\ref{app:matrix-2B}--\ref{app:matrix-8B} present the task-by-stage accuracy matrices. Rows indicate the most recently trained task, and columns indicate the evaluated task. Bold entries mark the diagonal; blank cells correspond to tasks not yet learned.

\newlength{\AppendixMatrixWidth}
\setlength{\AppendixMatrixWidth}{.49\linewidth}
\newcommand{\AppendixStageMatrix}[2]{%
\begin{minipage}[t]{\AppendixMatrixWidth}
\centering
\fontsize{9}{10.2}\selectfont
#1\par\vspace{1pt}
\setlength{\tabcolsep}{5pt}
\renewcommand{\arraystretch}{1.02}
\begin{tabular}{lrrrrr}
\toprule
& \rotatebox{50}{MedBook} & \rotatebox{50}{Navigation} & \rotatebox{50}{We-Math2} & \rotatebox{50}{Puzzle} & \rotatebox{50}{FinMME} \\
\midrule
#2
\bottomrule
\end{tabular}
\end{minipage}%
}
\par\smallskip
\begingroup\centering
\captionof{table}{Task accuracy (\%) after each training stage on Qwen3-VL-2B. Each panel shows one method.}
\label{app:matrix-2B}
\noindent\AppendixStageMatrix{(a) FullSFT}{
MedBook & \textbf{75.55} &  &  &  &  \\
Navigation & 71.31 & \textbf{77.92} &  &  &  \\
We-Math2 & 71.14 & 56.03 & \textbf{41.23} &  &  \\
Puzzle & 50.08 & 7.05 & 30.64 & \textbf{94.25} &  \\
FinMME & 45.67 & 19.11 & 33.90 & 84.00 & \textbf{32.69} \\
}\hfill%
\AppendixStageMatrix{(b) LoRA}{
MedBook & \textbf{74.36} &  &  &  &  \\
Navigation & 74.53 & \textbf{72.54} &  &  &  \\
We-Math2 & 70.97 & 63.82 & \textbf{39.76} &  &  \\
Puzzle & 63.67 & 37.11 & 36.27 & \textbf{92.25} &  \\
FinMME & 59.08 & 49.17 & 35.75 & 88.00 & \textbf{32.80} \\
}\par\smallskip
\noindent\AppendixStageMatrix{(c) O-LoRA}{
MedBook & \textbf{73.85} &  &  &  &  \\
Navigation & 72.67 & \textbf{56.22} &  &  &  \\
We-Math2 & 67.91 & 51.76 & \textbf{38.06} &  &  \\
Puzzle & 63.33 & 20.41 & 35.56 & \textbf{85.00} &  \\
FinMME & 57.72 & 22.82 & 35.22 & 75.50 & \textbf{35.17} \\
}\hfill%
\AppendixStageMatrix{(d) RegLoRA}{
MedBook & \textbf{73.34} &  &  &  &  \\
Navigation & 72.16 & \textbf{58.63} &  &  &  \\
We-Math2 & 71.31 & 56.59 & \textbf{37.77} &  &  \\
Puzzle & 66.04 & 41.93 & 35.84 & \textbf{86.25} &  \\
FinMME & 61.63 & 43.04 & 35.51 & 81.25 & \textbf{33.00} \\
}\par\smallskip
\noindent\AppendixStageMatrix{(e) GRPO}{
MedBook & \textbf{64.35} &  &  &  &  \\
Navigation & 63.50 & \textbf{39.52} &  &  &  \\
We-Math2 & 61.97 & 26.16 & \textbf{52.43} &  &  \\
Puzzle & 59.93 & 16.14 & 48.33 & \textbf{77.75} &  \\
FinMME & 50.93 & 19.11 & 45.14 & 69.50 & \textbf{41.78} \\
}\hfill%
\AppendixStageMatrix{(f) GRPO$^{*}$}{
MedBook & \textbf{57.89} &  &  &  &  \\
Navigation & 57.89 & \textbf{14.47} &  &  &  \\
We-Math2 & 58.06 & 16.14 & \textbf{51.45} &  &  \\
Puzzle & 56.37 & 18.18 & 49.96 & \textbf{64.25} &  \\
FinMME & 52.46 & 19.48 & 48.44 & 55.25 & \textbf{41.27} \\
}\par\smallskip
\noindent\AppendixStageMatrix{(g) GSPO}{
MedBook & \textbf{67.57} &  &  &  &  \\
Navigation & 65.03 & \textbf{66.05} &  &  &  \\
We-Math2 & 65.20 & 48.98 & \textbf{55.41} &  &  \\
Puzzle & 58.74 & 13.73 & 41.46 & \textbf{85.00} &  \\
FinMME & 47.03 & 13.36 & 40.24 & 76.75 & \textbf{43.14} \\
}\hfill%
\AppendixStageMatrix{(h) CPO}{
MedBook & \textbf{67.57} &  &  &  &  \\
Navigation & 66.21 & \textbf{63.64} &  &  &  \\
We-Math2 & 63.84 & 46.57 & \textbf{53.78} &  &  \\
Puzzle & 62.14 & 37.66 & 26.07 & \textbf{79.25} &  \\
FinMME & 56.54 & 33.58 & 47.34 & 75.75 & \textbf{42.33} \\
}\par\smallskip
\AppendixStageMatrix{\textbf{(i) CoLiR (Ours)}}{
\rowcolor{CoLiRTableTint}MedBook & \textbf{67.57} &  &  &  &  \\
\rowcolor{CoLiRTableTint}Navigation & 61.80 & \textbf{66.60} &  &  &  \\
\rowcolor{CoLiRTableTint}We-Math2 & 61.46 & 64.01 & \textbf{53.02} &  &  \\
\rowcolor{CoLiRTableTint}Puzzle & 60.61 & 59.55 & 49.65 & \textbf{88.75} &  \\
\rowcolor{CoLiRTableTint}FinMME & 53.14 & 52.69 & 45.80 & 82.50 & \textbf{42.48} \\
}\par\endgroup
\clearpage
\begingroup\centering
\captionof{table}{Task accuracy (\%) after each training stage on Qwen3-VL-4B. Each panel shows one method.}
\label{app:matrix-4B}
\noindent\AppendixStageMatrix{(a) FullSFT}{
MedBook & \textbf{77.76} &  &  &  &  \\
Navigation & 76.06 & \textbf{83.49} &  &  &  \\
We-Math2 & 73.85 & 66.05 & \textbf{49.06} &  &  \\
Puzzle & 53.31 & 58.26 & 40.26 & \textbf{86.00} &  \\
FinMME & 53.48 & 24.86 & 37.78 & 79.75 & \textbf{35.42} \\
}\hfill%
\AppendixStageMatrix{(b) LoRA}{
MedBook & \textbf{77.25} &  &  &  &  \\
Navigation & 74.87 & \textbf{60.11} &  &  &  \\
We-Math2 & 74.53 & 48.61 & \textbf{43.97} &  &  \\
Puzzle & 44.31 & 5.94 & 32.32 & \textbf{36.75} &  \\
FinMME & 66.89 & 45.64 & 42.26 & 35.00 & \textbf{41.62} \\
}\par\smallskip
\noindent\AppendixStageMatrix{(c) O-LoRA}{
MedBook & \textbf{81.32} &  &  &  &  \\
Navigation & 75.55 & \textbf{73.10} &  &  &  \\
We-Math2 & 78.44 & 59.37 & \textbf{46.12} &  &  \\
Puzzle & 76.06 & 54.36 & 45.24 & \textbf{93.00} &  \\
FinMME & 75.04 & 49.54 & 43.67 & 93.75 & \textbf{41.32} \\
}\hfill%
\AppendixStageMatrix{(d) RegLoRA}{
MedBook & \textbf{80.81} &  &  &  &  \\
Navigation & 71.65 & \textbf{82.93} &  &  &  \\
We-Math2 & 79.46 & 75.51 & \textbf{45.43} &  &  \\
Puzzle & 80.31 & 74.58 & 44.87 & \textbf{45.25} &  \\
FinMME & 76.91 & 74.95 & 44.70 & 37.75 & \textbf{41.02} \\
}\par\smallskip
\noindent\AppendixStageMatrix{(e) GRPO}{
MedBook & \textbf{71.65} &  &  &  &  \\
Navigation & 68.08 & \textbf{62.52} &  &  &  \\
We-Math2 & 68.59 & 43.78 & \textbf{58.08} &  &  \\
Puzzle & 61.46 & 32.84 & 48.62 & \textbf{77.00} &  \\
FinMME & 60.95 & 18.74 & 51.54 & 76.00 & \textbf{43.14} \\
}\hfill%
\AppendixStageMatrix{(f) GRPO$^{*}$}{
MedBook & \textbf{56.37} &  &  &  &  \\
Navigation & 50.25 & \textbf{1.48} &  &  &  \\
We-Math2 & 50.25 & 2.97 & \textbf{52.70} &  &  \\
Puzzle & 51.10 & 2.41 & 52.57 & \textbf{31.00} &  \\
FinMME & 48.90 & 2.04 & 53.18 & 28.25 & \textbf{41.17} \\
}\par\smallskip
\noindent\AppendixStageMatrix{(g) GSPO}{
MedBook & \textbf{73.85} &  &  &  &  \\
Navigation & 73.85 & \textbf{91.84} &  &  &  \\
We-Math2 & 71.31 & 84.97 & \textbf{64.02} &  &  \\
Puzzle & 68.08 & 32.84 & 54.16 & \textbf{92.00} &  \\
FinMME & 58.57 & 35.62 & 51.60 & 91.50 & \textbf{45.56} \\
}\hfill%
\AppendixStageMatrix{(h) CPO}{
MedBook & \textbf{73.85} &  &  &  &  \\
Navigation & 72.16 & \textbf{82.00} &  &  &  \\
We-Math2 & 71.99 & 79.22 & \textbf{65.24} &  &  \\
Puzzle & 69.44 & 69.76 & 59.08 & \textbf{83.50} &  \\
FinMME & 60.27 & 60.11 & 55.30 & 83.00 & \textbf{49.29} \\
}\par\smallskip
\AppendixStageMatrix{\textbf{(i) CoLiR (Ours)}}{
\rowcolor{CoLiRTableTint}MedBook & \textbf{73.85} &  &  &  &  \\
\rowcolor{CoLiRTableTint}Navigation & 68.93 & \textbf{87.01} &  &  &  \\
\rowcolor{CoLiRTableTint}We-Math2 & 69.27 & 91.28 & \textbf{63.26} &  &  \\
\rowcolor{CoLiRTableTint}Puzzle & 67.40 & 90.17 & 58.69 & \textbf{92.00} &  \\
\rowcolor{CoLiRTableTint}FinMME & 62.82 & 91.28 & 57.96 & 89.50 & \textbf{45.21} \\
}\par\endgroup
\clearpage
\begingroup\centering
\captionof{table}{Task accuracy (\%) after each training stage on Qwen3-VL-8B. Each panel shows one method.}
\label{app:matrix-8B}
\noindent\AppendixStageMatrix{(a) FullSFT}{
MedBook & \textbf{79.63} &  &  &  &  \\
Navigation & 79.12 & \textbf{86.46} &  &  &  \\
We-Math2 & 76.57 & 80.15 & \textbf{52.31} &  &  \\
Puzzle & 34.30 & 55.66 & 41.02 & \textbf{35.00} &  \\
FinMME & 58.57 & 25.60 & 43.20 & 35.75 & \textbf{37.79} \\
}\hfill%
\AppendixStageMatrix{(b) LoRA}{
MedBook & \textbf{82.34} &  &  &  &  \\
Navigation & 80.98 & \textbf{84.97} &  &  &  \\
We-Math2 & 79.12 & 75.70 & \textbf{49.91} &  &  \\
Puzzle & 47.37 & 21.89 & 26.15 & \textbf{94.75} &  \\
FinMME & 65.87 & 47.12 & 44.89 & 95.25 & \textbf{42.28} \\
}\par\smallskip
\noindent\AppendixStageMatrix{(c) O-LoRA}{
MedBook & \textbf{81.32} &  &  &  &  \\
Navigation & 79.97 & \textbf{62.34} &  &  &  \\
We-Math2 & 79.63 & 50.65 & \textbf{47.06} &  &  \\
Puzzle & 78.95 & 34.14 & 44.62 & \textbf{35.25} &  \\
FinMME & 72.50 & 52.32 & 44.78 & 35.25 & \textbf{42.79} \\
}\hfill%
\AppendixStageMatrix{(d) RegLoRA}{
MedBook & \textbf{82.68} &  &  &  &  \\
Navigation & 81.32 & \textbf{77.37} &  &  &  \\
We-Math2 & 78.78 & 74.03 & \textbf{47.88} &  &  \\
Puzzle & 77.08 & 74.03 & 46.53 & \textbf{62.00} &  \\
FinMME & 72.84 & 72.36 & 46.39 & 55.50 & \textbf{44.70} \\
}\par\smallskip
\noindent\AppendixStageMatrix{(e) GRPO}{
MedBook & \textbf{69.78} &  &  &  &  \\
Navigation & 71.99 & \textbf{12.62} &  &  &  \\
We-Math2 & 68.59 & 2.23 & \textbf{61.09} &  &  \\
Puzzle & 68.25 & 3.90 & 59.34 & \textbf{76.25} &  \\
FinMME & 62.14 & 3.90 & 56.35 & 67.00 & \textbf{45.16} \\
}\hfill%
\AppendixStageMatrix{(f) GRPO$^{*}$}{
MedBook & \textbf{71.31} &  &  &  &  \\
Navigation & 73.17 & \textbf{52.88} &  &  &  \\
We-Math2 & 71.99 & 34.69 & \textbf{64.28} &  &  \\
Puzzle & 70.97 & 31.35 & 63.20 & \textbf{62.25} &  \\
FinMME & 67.06 & 27.27 & 62.10 & 58.50 & \textbf{48.59} \\
}\par\smallskip
\noindent\AppendixStageMatrix{(g) GSPO}{
MedBook & \textbf{74.19} &  &  &  &  \\
Navigation & 72.67 & \textbf{88.68} &  &  &  \\
We-Math2 & 73.34 & 61.60 & \textbf{62.95} &  &  \\
Puzzle & 66.89 & 7.61 & 55.05 & \textbf{92.25} &  \\
FinMME & 58.74 & 3.15 & 56.37 & 90.25 & \textbf{45.46} \\
}\hfill%
\AppendixStageMatrix{(h) CPO}{
MedBook & \textbf{74.19} &  &  &  &  \\
Navigation & 73.34 & \textbf{85.34} &  &  &  \\
We-Math2 & 70.12 & 62.15 & \textbf{65.40} &  &  \\
Puzzle & 71.31 & 52.32 & 62.44 & \textbf{90.50} &  \\
FinMME & 64.52 & 33.58 & 56.43 & 89.25 & \textbf{47.12} \\
}\par\smallskip
\AppendixStageMatrix{\textbf{(i) CoLiR (Ours)}}{
\rowcolor{CoLiRTableTint}MedBook & \textbf{74.19} &  &  &  &  \\
\rowcolor{CoLiRTableTint}Navigation & 71.65 & \textbf{94.25} &  &  &  \\
\rowcolor{CoLiRTableTint}We-Math2 & 70.80 & 94.06 & \textbf{61.44} &  &  \\
\rowcolor{CoLiRTableTint}Puzzle & 67.06 & 91.47 & 58.03 & \textbf{87.25} &  \\
\rowcolor{CoLiRTableTint}FinMME & 60.61 & 94.81 & 56.37 & 84.00 & \textbf{45.06} \\
}\par\endgroup

\end{document}